\pdfoutput=1
\documentclass[11pt]{article}

\usepackage[final]{acl}

\usepackage{times}
\usepackage{latexsym}

\usepackage[T1]{fontenc}
\usepackage[utf8]{inputenc}

\usepackage{microtype}

\usepackage{inconsolata}

\usepackage{graphicx}

\usepackage{newfloat}
\usepackage{listings}
\usepackage{amsmath, amssymb}
\usepackage{multirow}
\usepackage{booktabs}
\usepackage{bbding}
\title{BrainECHO: Semantic Brain Signal Decoding through Vector-Quantized Spectrogram Reconstruction for Whisper-Enhanced Text Generation}

\author{
 \textbf{Jilong Li\textsuperscript{1}},
 \textbf{Zhenxi Song\textsuperscript{1}\thanks{Corresponding author}},
 \textbf{Jiaqi Wang\textsuperscript{1,2}},
 \textbf{Meishan Zhang\textsuperscript{1}},
\\
 \textbf{Honghai Liu\textsuperscript{1}},
 \textbf{Min Zhang\textsuperscript{1}},
 \textbf{Zhiguo Zhang\textsuperscript{1}\thanks{Corresponding author}}
\\
\\
 \textsuperscript{1}Harbin Institute of Technology, Shenzhen, China
\\
 \textsuperscript{2}Peng Cheng Laboratory, China
\\
 \small{
   \textbf{Correspondence:} \{songzhenxi, zhiguozhang\}@hit.edu.cn
 }
}

\begin{document}
\maketitle
\begin{abstract}

Current EEG/MEG-to-text decoding systems suffer from three key limitations: 
(1) reliance on teacher-forcing methods, which compromises robustness during inference, 
(2) sensitivity to session-specific noise, hindering generalization across subjects, and 
(3) misalignment between brain signals and linguistic representations due to pre-trained language model over-dominance.
To overcome these challenges, we propose BrainECHO (\textbf{B}rain signal decoding via v\textbf{E}ctor-quantized spe\textbf{C}trogram reconstruction for W\textbf{H}isper-enhanced text generati\textbf{O}n), a multi-stage framework that employs decoupled representation learning to achieve state-of-the-art performance on both EEG and MEG datasets. Specifically, BrainECHO consists of three stages: 
(1) Discrete autoencoding, which transforms continuous Mel spectrograms into a finite set of high-quality discrete representations for subsequent stages. 
(2) Frozen alignment, where brain signal embeddings are mapped to corresponding Mel spectrogram embeddings in a frozen latent space, effectively filtering session-specific noise through vector-quantized reconstruction, yielding a 3.65\% improvement in BLEU-4 score. 
(3) Constrained decoding fine-tuning, which leverages the pre-trained Whisper model for audio-to-text translation, balancing signal adaptation with knowledge preservation, and achieving 74\%-89\% decoding BLEU scores without excessive reliance on teacher forcing.
BrainECHO demonstrates robustness across sentence, session, and subject-independent conditions, passing Gaussian noise tests and showcasing its potential for enhancing language-based brain-computer interfaces.
% By decoupling brain signal representation learning from linguistic knowledge preservation, we mitigate the risk of LLM over-dominance and establish robust EEG-text mapping.

\end{abstract}

\section{Introduction}

Decoding text from brain activity, such as electroencephalography (EEG) and magnetoencephalography (MEG), is a critical and frontier research topic that can provide a foundation for language-based brain-computer interfaces (BCI) by enabling direct text input through brain signals. In the long term, accurate real-time translation of human brain signals can promote the widespread application of BCI technology in medicine, assistive technology, and entertainment, bringing new possibilities to human life.

With the rapid developments in natural language processing (NLP), automatic speech recognition (ASR), and other fields, researchers have leveraged the powerful language understanding and generating capabilities of pretrained large language models (LLMs) for neural decoding tasks~\citep{wang2022open, duan2024dewave, yang2024decode, yang2024mad}, making it possible to accurately decode text stimuli from non-invasive signals. EEG-to-Text~\citep{wang2022open} is the first work to decode open-vocabulary tokens from encoded word-level EEG rhythm features with the pretrained large model BART~\citep{lewis2020bart}. Furthermore, DeWave~\citep{duan2024dewave} used sentence-level raw EEG signals to perform EEG-to-text decoding without eye movement event markers. 

Later on, several BART-based methods~\citep{xi2023unicorn, feng2023aligning, amrani2024deep} were introduced, predominantly employing a pretraining-finetuning paradigm. These methods first align EEG representations with pretrained text embeddings before feeding them into BART for finetuning. Although these approaches have yielded impressive results, they rely on a teacher-forcing generation strategy, wherein the model depends on the ground truth preceding text during each token prediction. This setting does not accurately reflect the model's performance in real-world scenarios. These methods show poor decoding performance without teacher forcing.

To address this limitation, NeuSpeech~\citep{yang2024decode} and MAD~\citep{yang2024mad} treat raw MEG signals as a specialized form of speech, transforming MEG signals and feeding them into a pre-trained Whisper model~\citep{radford2023robust}, which is trained on large-scale audio-text pairs, for end-to-end text decoding without teacher forcing. However, these approaches primarily focus on mapping continuous brain signals to discrete text without compressing the signals into discrete representations, thereby limiting the model's decoding accuracy and generalization capabilities.
% Extensive researches in speech recognition~\citep{zhang2023dub, puvvada2024discrete} demonstrate that discrete representations preserve more semantic information for translation compared to conventional speech features like Fbank, thanks to their carefully designed self-supervised learning paradigms. 
% While DeWave~\citep{duan2024dewave} aligns discrete representations of input EEG signals and text, it assumes a chronological order for the discrete token sequence, requiring a highly capable feature extractor. Considering the natural temporal alignment between audio-evoked brain signals and audio stimuli, aligning raw signals and speech within a discrete space leverages implicit temporal properties, thereby reducing the difficulty of converting neural signals into human language. 

\begin{figure}[t]
\centering
\includegraphics[width=0.48\textwidth]{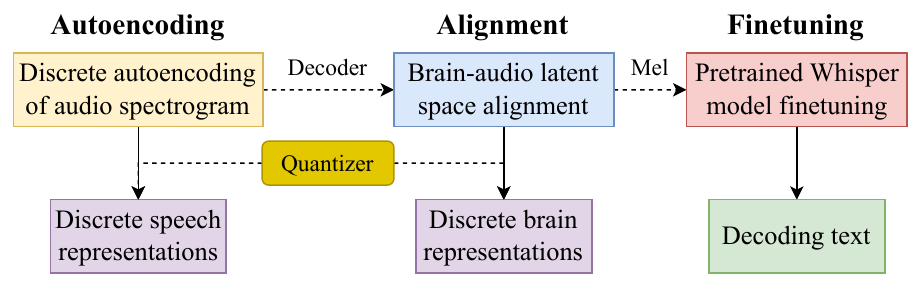}
\caption{
Overview of the BrainECHO framework learning process, illustrated through a simplified conceptual diagram for enhanced understanding.
% Learning process overview of our proposed BrainECHO framework.
BrainECHO follows a three-stage \textit{autoencoding–alignment–finetuning} paradigm to achieve decoupled representation learning:
\textbf{\textit{Autoencoding}} stage is used to warm up the Mel spectrogram reconstruction by employing a codebook-based quantizer to enhance generalizability and robustness. This stage especially focuses on exploiting discrete representations.
\textbf{\textit{Alignment}} stage reconstructs the Mel spectrogram from the corresponding aurally evoked brain signals. This involves designing a new brain encoder that integrates with the warmed-up quantizer and decoder from the first stage.
\textbf{\textit{Finetuning}} stage leverages the capabilities of the pre-trained Whisper model to achieve audio-text translation.
% Revised/ZS
}
\label{fig:process}
\end{figure}

% In brain-to-text decoding, introducing discrete representation helps solve two fundamental challenges. First, EEG/MEG signals are inherently contaminated by physiological artifacts (e.g., muscle movements, ocular noise) and session-specific variability (e.g., electrode impedance shifts). By discretizing brain signals into a finite codebook, vector quantization acts as a sparsity-inducing filter that discards high-frequency, unstructured noise while preserving low-frequency semantic patterns~\citep{duan2024dewave}. Second, the codebook's discrete latent space serves as a modality-invariant interface, enabling seamless alignment between brain signals and text tokens. This avoids the "distribution shift" problem in end-to-end continuous-to-discrete mapping, which can lead to spurious correlations~\citep{shirakawa2024spurious}.
In brain-to-text decoding, introducing discrete representation helps solve two fundamental challenges. First, the codebook's discrete latent space serves as a modality-invariant interface, enabling seamless alignment between brain signals and text tokens. This avoids the "distribution shift" problem in end-to-end continuous-to-discrete mapping, which can lead to spurious correlations~\citep{shirakawa2024spurious}. Second, EEG/MEG signals are inherently contaminated by physiological artifacts (e.g., muscle movements, ocular noise) and session-specific variability (e.g., electrode impedance shifts). By discretizing brain signals into a semantic-pruned codebook, vector quantization acts as a sparsity-inducing filter that discards task-irrelevant embeddings. This mechanism is analogous to noise suppression in VQ-VAE-based models~\citep{razavi2019generating}.
 
Therefore, we propose a novel multi-stage semantic decoding framework for EEG/MEG \textbf{brain} signals, aurally evoked by semantic audio, through v\textbf{E}ctor-quantized spe\textbf{C}trogram reconstruction for W\textbf{H}isper-enhanced text generati\textbf{O}n, termed \textbf{BrainECHO}. % -- Revised/ZS
% Specifically, BrainECHO executes the following steps: % --Added/ZS
% 1) Discrete \textit{autoencoding} of the audio spectrogram, particularly employing codebook-based vector quantization, to establish a pre-warmed representation space that facilitates Mel spectrogram reconstruction; % -- Revised/ZS
% 2) Brain-audio latent space \textit{alignment}, utilizing a brain encoder and pre-warmed quantizer and decoder to reconstruct the evoked brain signal’s Mel spectrogram; % -- Revised/ZS
% 3) Semantic text generation, achieved through AdaLoRA-based \textit{finetuning} of the pre-trained Whisper model, with the reconstructed Mel spectrogram as input. % -- Revised/ZS
The overall three-stage (\textit{autoencoding, alignment, finetuning}) training process of the proposed BrainECHO is illustrated in Figure~\ref{fig:process}. % --Added/ZS
We validate the performance of BrainECHO using two different public audio-evoked brain signal datasets: \textit{Brennan}, which contains EEG data, and \textit{GWilliams}, which contains MEG data.
The principal contributions of our work are summarized below:
\begin{itemize}

\item The proposed BrainECHO framework addresses EEG/MEG-to-text limitations of teacher-forcing dependency and poor Gaussian noise generalization~\citep{jo2024eeg, wang2022open}, achieving SOTA performance on EEG and MEG benchmarks~\citep{brennan2019hierarchical, gwilliams2023introducing}. Its robustness is further validated through novel subject/session-independent data splits, addressing a critical gap in prior research.

\item Unlike recent non-teacher-forcing methods~\citep{yang2024decode, yang2024mad} that directly fine-tune LLMs, BrainECHO mitigates LLM overfitting risks through a multi-stage training strategy, effectively balancing noise suppression in brain signals with preservation of pre-trained linguistic knowledge.

\item By introducing a quantized codebook for discrete brain-signal representation---contrary to continuous latent spaces in prior work---BrainECHO filters session-specific noise and captures subject-invariant semantics, achieving SOTA cross-subject generalization.

\end{itemize}

\section{Related Works}
Non-invasive brain signals such as EEG and MEG offer significant advantages over invasive alternatives, particularly in terms of safety and cost-effectiveness. Considerable progress has been made in decoding text from noninvasive signals.
% \subsection{Closed-Vocabulary Neural Decoding}
Ghazaryan et al.~\citep{ghazaryan2023trials} utilized Word2vec to decode 60 nouns from MEG recordings. Meta~\citep{defossez2023decoding} developed a model that uses wav2vec 2.0~\citep{baevski2020wav2vec} and contrastive learning to decode speech from 3-second EEG/MEG signals. However, these methods are restricted to decoding a small set of words or segments, restricting their applicability to open-vocabulary text generation.

% \subsection{Decoder-Only Architectures for Open-Vocabulary Brain-to-Text Decoding}
\subsection{Decoder-Only Models for Brain-to-Text}
Recent advancements have leveraged the powerful understanding and generation capabilities of pretrained models, particularly LLMs, to extend vocabulary from closed to open. In decoder-only architectures, some researchers have aligned brain signals with text to guide pretrained generative models in text generation. For example, Tang et al. \citep{tang2023semantic} and Zhao et al. \citep{zhao2024mapguide} mapped fMRI data to text embeddings to iteratively guide GPT-2 in generating text. Similarly, Chen et al. \citep{chen2024open} used text-aligned fMRI representations as prompts for GPT-2 to decode language information. 

% \subsection{Seq2seq Architectures for Open-Vocabulary Brain-to-Text Decoding}
\subsection{Seq2Seq Models for Brain-to-Text}
Wang et al.\citep{wang2022open} fed transformed word-level EEG rhythm feature into a pretrained BART model to decode open-vocabulary tokens. Duan et al.\citep{duan2024dewave} integrated discrete EEG encodings with text-EEG contrastive alignment to mitigate individual variability in brain activity. However, these BART-based methods rely on teacher forcing during inference. Furthermore, as Jo et al.~\citep{jo2024eeg} demonstrated, their performance on noisy data is comparable to that on EEG data, suggesting that these models may simply memorize the training data. Recently, NeuSpeech~\citep{yang2024decode} directly fed raw MEG signals into a modified, pretrained Whisper model for text decoding without teacher forcing. Furthermore, MAD~\citep{yang2024mad} introduced MEG-speech alignment loss to decode sentences not present in the training data. However, these Whisper-based methods do not utilize discrete representations of the original signals to enhance the model's generalization capabilities. Our work integrates brain-audio discretization and alignment, aiming to predict high-quality Mel spectrograms from brain signals that align with Whisper's input format. Leveraging Whisper's advanced speech recognition abilities, our approach generates sentences that closely mirror the original text.

\begin{figure*}[t]
\centering
\includegraphics[width=1.0\textwidth]{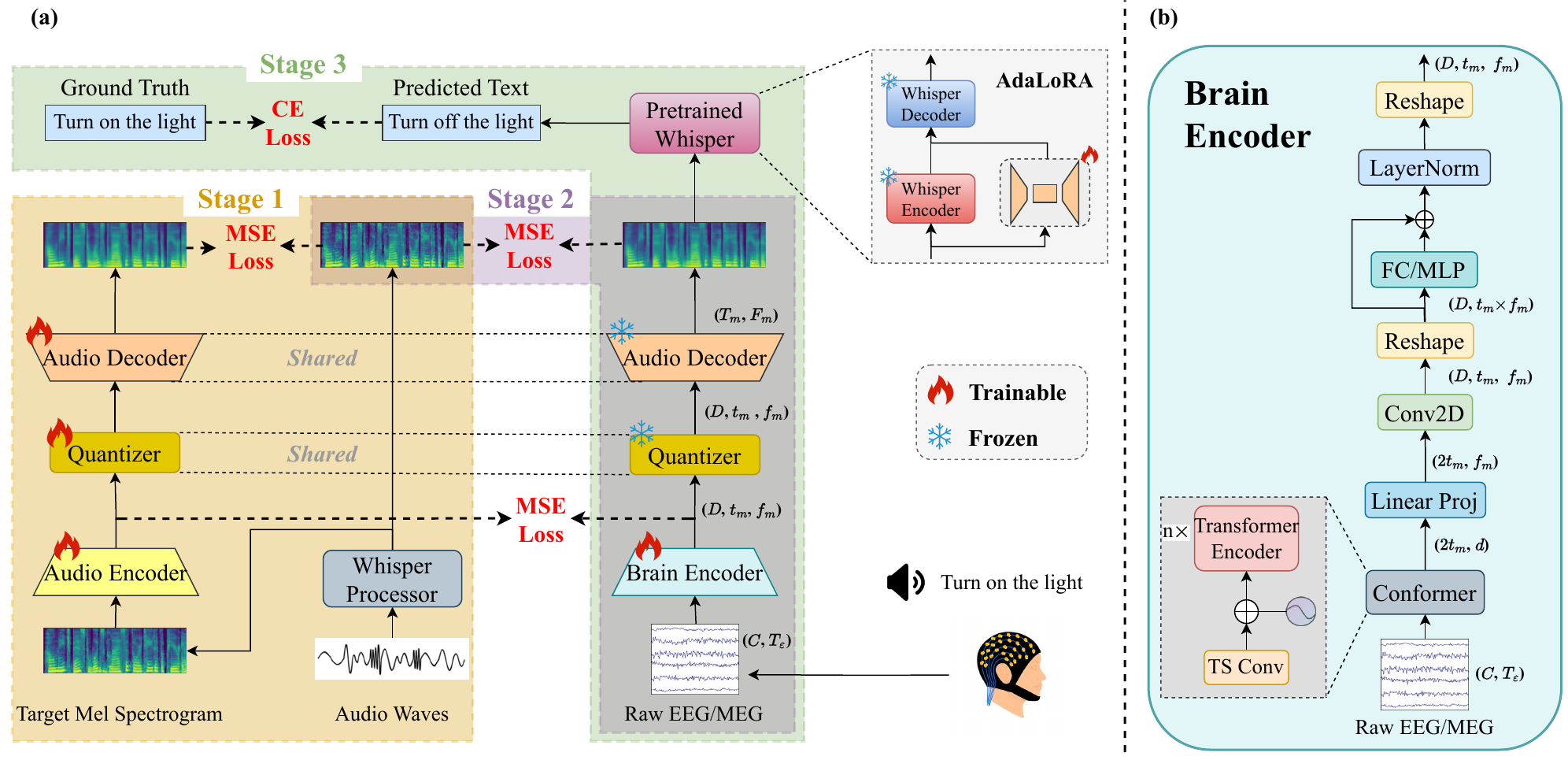}
\caption{(a) Overview of the BrainECHO model framework. BrainECHO utilizes a three-stage training paradigm consisting of Mel spectrogram autoencoding, brain-audio latent space alignment, and Whisper finetuning. $C$, $T_{\varepsilon }$ denotes numbers of raw wave channels and timestamps, respectively. (b) Details of the Brain Encoder, which converts raw EEG/MEG signals into latent representations. $d$ represents the dimension of hidden states and TS Conv stands for Spatio-Temporal Convolution Networks. More details of Conformer are provided in Appendix~\ref{sec:conformer}.}
\label{fig:model}
\end{figure*}

\section{Method}
\subsection{Task Definition}

Given the raw EEG/MEG $E$, text content $T$, and corresponding audio stimuli $A$ during listening as mentioned in Section~\ref{sec:dataset}, the experimental data can be divided into a series of sentence-level EEG/MEG-text-speech pairs $\left \langle \varepsilon ,t,a \right \rangle$. $\varepsilon \in \mathbb{R} ^{C_{\varepsilon }\times T_{\varepsilon }}$, where $C_{\varepsilon }$ and $T_{\varepsilon }$ represent the channels and timestamps of brain signals, respectively. In general, $T_{\varepsilon }$ varies with the length of the sentence-level audio segment. Our goal is to decode the corresponding open-vocabulary tokens $t$ from the brain signal $\varepsilon$, with $a$ serving as auxiliary information.

% \subsection{Model}
% \subsection{BrainECHO} 
\subsection{Model Architecture}

% The overall structure of our proposed model is illustrated in Figure~\ref{fig:process}. 
Unlike the multi-task joint training employed in MAD~\citep{yang2024mad}, BrainECHO adopts a three-stage training process. This method reduces resource consumption at each training step and facilitates the prediction of high-quality, high-resolution Mel spectrograms from brain signals. Specifically, we extend the spectrogram duration from 3 seconds, as used in ~\citep{defossez2023decoding, yang2024mad}, to over 10 seconds, enabling sentence-level rather than segment-level brain-to-text translation, thereby preserving the semantics of the original sentences. The details of the model are shown in Figure~\ref{fig:model}. The following sections will detail each training stage.

% \subsubsection{Discrete Autoencoding of Audio Spectrogram}
\subsubsection{Autoencoding of Audio Spectrogram}

Van den Oord et al. introduced the Vector Quantized-Variational AutoEncoder (VQ-VAE)~\citep{van2017neural} to learn discrete latent representations of audio, video, and other data types. Building on this approach, several studies~\citep{li2023textless, sadok2023vector, yang2023diffsound} have explored representing Mel spectrograms using discrete tokens to capture phoneme-like information.
% Inspired by these methods, our first stage involves autoencoding Mel spectrograms, with the purpose of obtaining a discrete representation space that is conducive to Mel reconstruction. 
Since Mel spectrograms effectively capture frequency and temporal patterns of audio, it is feasible to use them as an intermediate modality between brain signals and text~\citep{metzger2023high, defossez2023decoding}. Due to the fact that the majority of existing strong audio autoencoders are pre-trained on audio waves rather than Mel, we chose to autoencode Mel spectrograms for obtaining a discrete representation space that is conducive to Mel reconstruction.
Specifically, given a spectrogram $m \in \mathbb{R} ^{T_{m}\times F_{m}}$, the audio encoder $Enc$ first converts it into a feature map $z_{m}=Enc(m) \in \mathbb{R} ^{t_{m}\times f_{m}\times D}$ , where $T_{m}$, $F_{m}$ and $D$ denote the number of time frames, frequency bins and latent channels, respectively. The spectrogram is generated by the Whisper Processor, enabling text decoding from the reconstructed spectrogram using Whisper's encoder-decoder architecture. Then, $z_m$ is processed by a vector quantizer $Q$. Specifically, each latent embedding $z_m^{ij} \in \mathbb{R} ^D$ ($1\le i\le t_{m}, 1\le j\le f_{m}$) is replaced by the nearset vector $z_q^{ij}$ from a codebook $\mathbb{C} \in \mathbb{R} ^{N \times D}$, which consists of $N$ learnable D-dimensional vectors. Formally, this process is expressed as follows:
\begin{equation}
\begin{split}
Q(z_m^{ij})&=z_q^{ij} = c_k, \\ 
\text{ where }k &= \mathop{\arg\min}\limits_{k \in \{1,2,...,N\}} \left \| z_m^{ij}-c_k \right \|_2 .
\end{split}
\end{equation}
The reconstructed spectrogram is then obtained by the audio decoder $Dec$ as: $\hat{m}=Dec(z_q)$.
The encoder and decoder are both composed of ResUNet blocks~\citep{kong2021decoupling}. 
% Our objective is to ensure the reconstructed spectrogram closely matches the target and compress the rich audio semantic information into the codebook. 
The training objective at this stage is defined as follows:
\begin{equation}
\begin{split}
    L_1&=\left \| m-\hat{m} \right \|_2^2 + \alpha \left \| sg(z_m)-z_q \right \|_2^2 \\
    &+ \beta_1 \left \| z_m-sg(z_q) \right \|_2^2,
\end{split}
\end{equation}
where $sg(\cdot)$ is a function for stopping gradients, and $\alpha$, $\beta_1$ are hyperparameters for the quantization loss and commitment loss weights, respectively.

% \subsubsection{Brain-Speech Modality Alignment}
\subsubsection{Brain-Audio Latent Space Alignment}

In the second stage, we freeze all the modules pre-trained in the previous stage and train a brain encoder to convert raw EEG/MEG signals $\varepsilon$ into latent features $z_{\varepsilon}$. The brain encoder utilizes a Conformer-based architecture~\citep{song2022eeg}, which begins with Spatio-Temporal Convolutional Networks to process the input signals into a one-dimensional embedding sequence. The spatial convolutional layer reduces the number of input signal channels to one, while the temporal convolutional layers downsample the time dimension. This sequence is then added to learnable position embeddings and fed into a stack of Transformer encoder blocks. Linear layers and 2D convolutional networks subsequently transform the EEG/MEG features into representations matching the shape of $z_m$. Similarly, $z_{\varepsilon}$ is input into the frozen quantizer $Q$ and audio decoder $Dec$ to predict the corresponding Mel spectrogram $m$. Additionally, we align the representations of the Mel spectrogram and raw signals in the latent space. Notably, we employ a unified codebook to leverage pre-warmed discrete acoustic tokens for representing brain activity. 
% The introduction of vector quantization enhances the stability and generalization of the Mel spectrogram reconstruction from brain signals, thereby improving the performance of subsequent text decoding. 
Formally, the loss for stage 2 is as follows:
\begin{equation}
\begin{split}
    L_2&=\left \| m-Dec(Q(z_{\varepsilon})) \right \|_2^2 + \gamma \left \| z_m-z_{\varepsilon} \right \|_2^2 \\
    &+ \beta_2 \left \| z_{\varepsilon}-sg(Q(z_{\varepsilon})) \right \|_2^2,
\end{split}
\end{equation}
where $\gamma$ and $\beta_2$ are used to scale the latent alignment loss and the commitment loss, respectively. The intermediate representations of the codebook and speech provide additional supervisory signals to guide the generation of Mel spectrograms. We employ L2 loss rather than CLIP loss~\citep{defossez2023decoding, yang2024mad} to generate highly restored spectrograms that match Whisper's input.

\subsubsection{Whisper Finetuning}
After obtaining the predicted Mel spectrogram, it is fed into the pretrained Whisper-base\footnote{\url{https://huggingface.co/openai/whisper-base.en}} model to decode tokens. Guided by both the need for computational efficiency and the proven success of this method in related work~\citep{yang2024decode, yang2024mad}, we utilize AdaLoRA~\citep{zhang2023adalora} to fine-tune its encoder while keeping the remaining parameters frozen. The objective is to minimize the cross-entropy loss between the predicted sentence and the ground truth $t$. While it is feasible to integrate the previous stages and this stage into one stage for end-to-end training, we adopt a three-stage framework for decoupled representation learning and training cost reduction. More discussion is presented in Appendix~\ref{sec:necessity}.
% \begin{equation}
% \begin{split}
%     L_3&=-\sum_{n=1}^{N}log\:p (t_n),
% \end{split}
% \end{equation}
% where $N$ denotes the number of tokens in the ground truth sentence.

\begin{table*}[t]
\small
\centering
\resizebox{\textwidth}{!}{
\begin{tabular}{lllcccccccc}
\toprule
\multicolumn{3}{c}{} & \multicolumn{4}{c}{\textbf{BLEU-N (\%) $\uparrow$}} & \multicolumn{3}{c}{\textbf{ROUGE-1 (\%)}$\uparrow$} & \multicolumn{1}{c}{\textbf{WER (\%) $\downarrow$}} \\
 \cmidrule(lr){4-7} \cmidrule(lr){8-10} \cmidrule(lr){11-11}
\textbf{Split} & \textbf{Input} & \textbf{Method} & N=1 & N=2 & N=3 & N=4 & P  & R & F & \\
\midrule
Subject & Noise & NeuSpeech~\cite{yang2024decode} & 8.45 & 1.78 & 0.43 & 0 & 10.26 & 21.61 & 13.02 & 198.31 \\
 & Noise & BrainECHO & 4.75 & 1.10 & 0.28 & 0 & 11.25 & 7.81 & 8.52 & 105.27 \\
 \cmidrule(lr){2-11}
 & EEG feature & EEG-to-Text~\cite{wang2022open} & 8.82 & 3.15 & 1.90 & 1.44 & 10.13 & 21.61 & 13.12 & 233.99 \\
 & EEG & NeuSpeech~\cite{yang2024decode} & 85.31 & 84.38 & 83.98 & 83.75 & 82.60 & 82.73 & 82.64 & 16.97 \\ 
 & EEG & MAD~\citep{yang2024mad} & 80.34 & 79.10 & 78.46 & 78.15 & 81.00 & \textbf{90.76} & 83.79 & 42.14 \\ 
 & EEG & BrainECHO & \textbf{89.78} & \textbf{89.06} & \textbf{88.74} & \textbf{88.55} & \textbf{87.05} & 87.27 & \textbf{87.13} & \textbf{11.72} \\ 
 \cmidrule(lr){2-11}
% EEG & NeuSpeech~\cite{yang2024decode} \textit{w/ tf} & 93.77 & 93.51 & 93.25 & 92.98 & 95.41 & 98.09 & 96.43 & 6.55 \\
 & EEG & BrainECHO \textit{w/ tf} & \textbf{98.82} & \textbf{98.74} & \textbf{98.68} & \textbf{98.64} & \textbf{98.45} & \textbf{98.44} & \textbf{98.45} & \textbf{1.18} \\
\midrule
Sentence & EEG & BrainECHO & \textbf{89.24} & \textbf{88.52} & \textbf{88.18} & \textbf{88.01} & \textbf{85.56} & \textbf{85.78} & \textbf{85.63} & \textbf{12.34} \\

\bottomrule
\end{tabular}
}
\centering
\caption{\label{tab:evaluation_results} Overall comparison of decoding performance on the \textit{Brennan} dataset.
% By default, all methods are evaluated without teacher forcing.
% The metrics with teacher forcing (\textit{w/ tf}) are further explored. 
% Additional evaluation results are presented in the Appendix.
% Further results and discussions are provided in Appendix~\ref{sec:eval}.
}
\end{table*}

\begin{table*}[t]
\small
\resizebox{\textwidth}{!}{
\centering
\begin{tabular}{lllcccccccc}
\toprule
\multicolumn{3}{c}{} & \multicolumn{4}{c}{\textbf{BLEU-N (\%) $\uparrow$}} & \multicolumn{3}{c}{\textbf{ROUGE-1 (\%)}$\uparrow$} & \multicolumn{1}{c}{\textbf{WER (\%) $\downarrow$}} \\
 \cmidrule(lr){4-7} \cmidrule(lr){8-10} \cmidrule(lr){11-11}
\textbf{Split} & \textbf{Input} & \textbf{Method} & N=1 & N=2 & N=3 & N=4 & P  & R & F & \\
\midrule
Random & MEG feature & EEG-to-Text~\cite{wang2022open} & 9.21 & 2.13 & 0.57 & 0.14 & 9.74 & 10.73 & 11.38 & 118.25 \\
Shuffling & MEG & NeuSpeech~\cite{yang2024decode} & 50.49 & 46.85 & 44.42 & 42.55 & 46.39 & 52.48 & 47.10 & 71.17 \\
 & MEG & NeuSpeech (Original results) & 60.3 & 55.26 & 51.24 & 47.78 & 60.88 & 59.76 & 58.73 & 56.63 \\
 & MEG & MAD~\cite{yang2024mad} & 3.93 & 0.42 & 0 & 0 & 8.98 & 6.85 & 7.26 & 105.33 \\
 & MEG & BrainECHO & \textbf{73.35} & \textbf{72.66} & \textbf{72.46} & \textbf{72.42} & \textbf{69.66} & \textbf{70.12} & \textbf{69.73} & \textbf{31.44} \\ 
\midrule
Session & MEG & NeuSpeech~\cite{yang2024decode} & 53.16 & - & - & - & - & - & - & - \\ 
 & MEG & BrainECHO & \textbf{75.24} & \textbf{74.57} & \textbf{74.34} & \textbf{74.27} & \textbf{72.94} & \textbf{72.84} & \textbf{72.78} & \textbf{29.59} \\ 
\midrule
Sentence & MEG & BrainECHO & \textbf{73.58} & \textbf{72.99} & \textbf{72.82} & \textbf{72.79} & \textbf{70.38} & \textbf{70.75} & \textbf{70.73} & \textbf{31.11} \\ 
\midrule
Subject & MEG & BrainECHO & \textbf{75.05} & \textbf{74.38} & \textbf{74.18} & \textbf{74.14} & \textbf{71.83} & \textbf{72.02} & \textbf{71.72} & \textbf{29.80} \\ 
\bottomrule
\end{tabular}
}
\centering
\caption{\label{tab:supp_eval_results} Overall comparison of decoding performance on the \textit{GWilliams} dataset. 
% All methods are evaluated without teacher forcing.
}
\end{table*}

\section{Experiments}
\subsection{Dataset}
\label{sec:dataset}
The \textit{Brennan} dataset~\citep{brennan2019hierarchical} comprises 49 human EEG recordings, of which 33 remained after screening. Participants passively listened to a 12.4-minute audiobook recording while their EEG signals were recorded. The \textit{GWilliams}~\citep{gwilliams2023introducing} dataset contains raw MEG recordings from 27 English speakers who listened to naturalistic stories for 2 hours. More details are provided in Appendix~\ref{sec:datasets}.

\subsection{Preprocessing}  
Brain signals in both datasets are preprocessed similarly. The EEG signals are notch-filtered at 60 Hz and bandpass-filtered between 0.5 and 99 Hz, and then resampled to 200 Hz. The MEG signals are notched at 50 Hz, filtered with 1$\sim$58 Hz and resampled to 100 Hz. Both datasets are normalized to a range of -1 to 1 using robust scalar.

All audio is resampled to 16,000 Hz to align with Whisper's pretraining configuration. To assess the robustness of our proposed method, we employ different approaches to generate samples. For the \textit{Brennan} dataset, we utilize WhisperX~\citep{bain2023whisperx}, a time-accurate speech recognition system, to segment the audio into chunks of up to 12 seconds. For the \textit{GWilliams} dataset, we split the audio according to the original annotations, resulting in segments no longer than 24 seconds. This process generates a series of EEG/MEG-text-speech pairs.

The Whisper processor then converts the speech into an 80-channel Mel spectrogram $m$ using 25-ms windows with a stride of 10 ms. To standardize settings and reduce memory usage, the length of the Mel spectrograms in \textit{GWilliams} is downsampled to half its original value, resulting in $m$ having a consistent shape of (80, 1200). Finally, we obtain 140 and 661 unique sentences from the two datasets, respectively. 

\subsection{Dataset Splitting and Validation Strategies}
Individual differences and attention levels of subjects can affect EEG signals, making it difficult for models to generalize across subjects and trials. To explore the model's generalization ability, we design different dataset splitting and validation strategies: random shuffling, session-based, sentence-based, and subject-based splittings. More details about the splitting strategies are provided in Appendix~\ref{sec:split}. We ensure that the test data are completely separate from the training data. However, we must note that our data split is done in pairs, i.e., "Semantic Audio -- Brain Signal evoked by the Semantic Audio." This means that the same sentence, evoking the same brain signal in a specific trial, will not appear in both the training and testing stages. Unless otherwise specified, the \textit{Brennan} and \textit{GWilliams} datasets are partitioned by subject-based splittings and random shuffling, respectively, in the following results.

\subsection{Implementation Details}
The models are trained on Nvidia 3090 GPUs (24GB). Training on the \textit{Brennan} and \textit{GWilliams} datasets take approximately 4 and 24 hours, respectively, using a single GPU. The hyperparameters are set as follows: $\alpha=0.5$, $\beta_1=\beta_2=0.1$, $\gamma=1$, $N=2048$, $d=256$, and $D=8$. The audio encoder is configured with a downsampling rate of 4. We use a vanilla Transformer encoder with 4 layers and 8 heads. All EEG/MEG samples are zero-padded to 2400 in the time dimension. Input spectrograms are padded uniformly to a length of 3000 with -1 following Whisper's configuration. For the \textit{GWilliams} dataset, the length of the predicted Mel spectrogram is upsampled by a factor of 2. When generating with Whisper, we set the number of beams to 5 for beam search and apply a repetition penalty of 5.0 with a no-repeat n-gram size of 2. Further details on the training configuration are provided in Appendix~\ref{sec:imp}.
% reducing the width and height of the input Mel spectrogram to one-fourth of their original size.

% \subsection{Evaluation Metrics}

% \subsection{Experimental Results}
\subsection{Comparative Study}
\label{sec:results}

% \textcolor{red}{We use BLEU~\citep{papineni2002bleu}, ROUGE-1~\citep{lin2004rouge} and Word Error Rate (WER) to evaluate decoding performance. BLEU and ROUGE-1 are used to evaluate the quality of text generation, while WER is used to calculate error rates based on edit distance.}

We use BLEU~\citep{papineni2002bleu}, ROUGE-1~\citep{lin2004rouge}, and Word Error Rate (WER) to evaluate decoding performance. BLEU and ROUGE-1 assess the quality of text generation, while WER calculates error rates based on edit distance.

% \subsubsection{Overall Comparison}
\subsubsection{Benchmarking SOTA Methods on the \textit{Brennan} Dataset}
We compare our model with popularly-referred brain-to-text architectures, i.e., EEG-to-Text~\citep{wang2022open}, NeuSpeech~\citep{yang2024decode} and MAD~\citep{yang2024mad}. NeuSpeech~\cite{yang2024decode}, the previous SOTA model for MEG-to-text translation, serves as the baseline for comparison. MAD~\citep{yang2024mad} introduces brain-audio alignment on the basis of NeuSpeech. To ensure a fair comparison, we replicated these frameworks based on our data split settings, using the same training and test data as input.
As shown in Table~\ref{tab:evaluation_results}, our method demonstrates remarkable decoding performance, achieving BLEU-\{1, 2, 3, 4\} of 89.78, 89.06, 88.74 and 88.55, as well as WER of 11.27 without teacher forcing. The results indicate that BrainECHO generates text highly consistent with the ground truth. Specifically, in terms of BLEU-4, BrainECHO outperforms the previous baseline and current SOTA method by 87.11 (+6049\%) and 4.8 (+5.73\%) respectively. When using teacher forcing, BrainECHO achieves BLEU-4 of 98.45, which is nearly perfect, highlighting the unrealistic metrics produced by teacher forcing evaluation.

Additionally, since BrainECHO is a generative model, it always produces some output, even with noise, which occasionally matches a few words and gets non-zero BLEU scores. However, the BLEU-4 score of zero shows that matching four consecutive words is unlikely. The noise results are still far from the EEG/MEG results, 
indicating that BrainECHO captures the intrinsic connection between brain signals and text, rather than simply memorizing sentences from the training set. Intuitively, BrainECHO is more resistant to noise than NeuSpeech~\citep{yang2024decode}. Notably, the model ideally should not respond to noise, with a WER expected to be 1. Therefore, a high WER ($>$ 1), suggesting the model outputs excessive irrelevant content, is not necessarily a desirable result.

% \subsubsection{Evaluation Results}
\subsubsection{Benchmarking SOTA Methods on the \textit{GWilliams} Dataset}
\label{sec:eval}
Evaluation metrics on the \textit{GWilliams} dataset across various splitting strategies are presented in Table~\ref{tab:supp_eval_results}. 
When using random shuffling, BrainECHO achieves a BLEU-4 score of 72.42, outperforming NeuSpeech by 24.64 points (+51.57\%). Furthermore, with session-based division, BrainECHO achieves a BLEU-1 score of 75.24, exceeding NeuSpeech by 22.08 points (+41.53\%). These results indicate that BrainECHO can generate text that closely matches the ground truth. Additionally, the results we reproduced on MAD are unsatisfactory on both datasets, especially on GWilliams, indicating that optimizing the CLIP loss between neural signals and audio representations is particularly challenging when the input signal is long (the original experimental setup in MAD used only a 4-second time length). 
Some examples of generated sentences are presented in Appendix~\ref{sec:samples}.

Additionally, the performances across various splitting strategies are presented. BrainECHO demonstrates optimal performance on the \textit{GWilliams} dataset when split by sessions. In particular, the performance differences are not significant, indicating that BrainECHO is robust and effectively alleviates covariate shift among different subjects or trials without the need for external information (e.g., subject or trial identifiers), provided that all unique sentences are encountered during training. In contrast, the brain module used in~\citep{defossez2023decoding, yang2024mad} employs distinct projection matrices for each subject to mitigate individual differences, yet it cannot be generalized to unseen subjects directly. More discussion about the rationality of splitting strategies is provided in Appendix~\ref{sec:sense}.

\begin{figure}[t]
\centering
\includegraphics[width=0.50\textwidth]{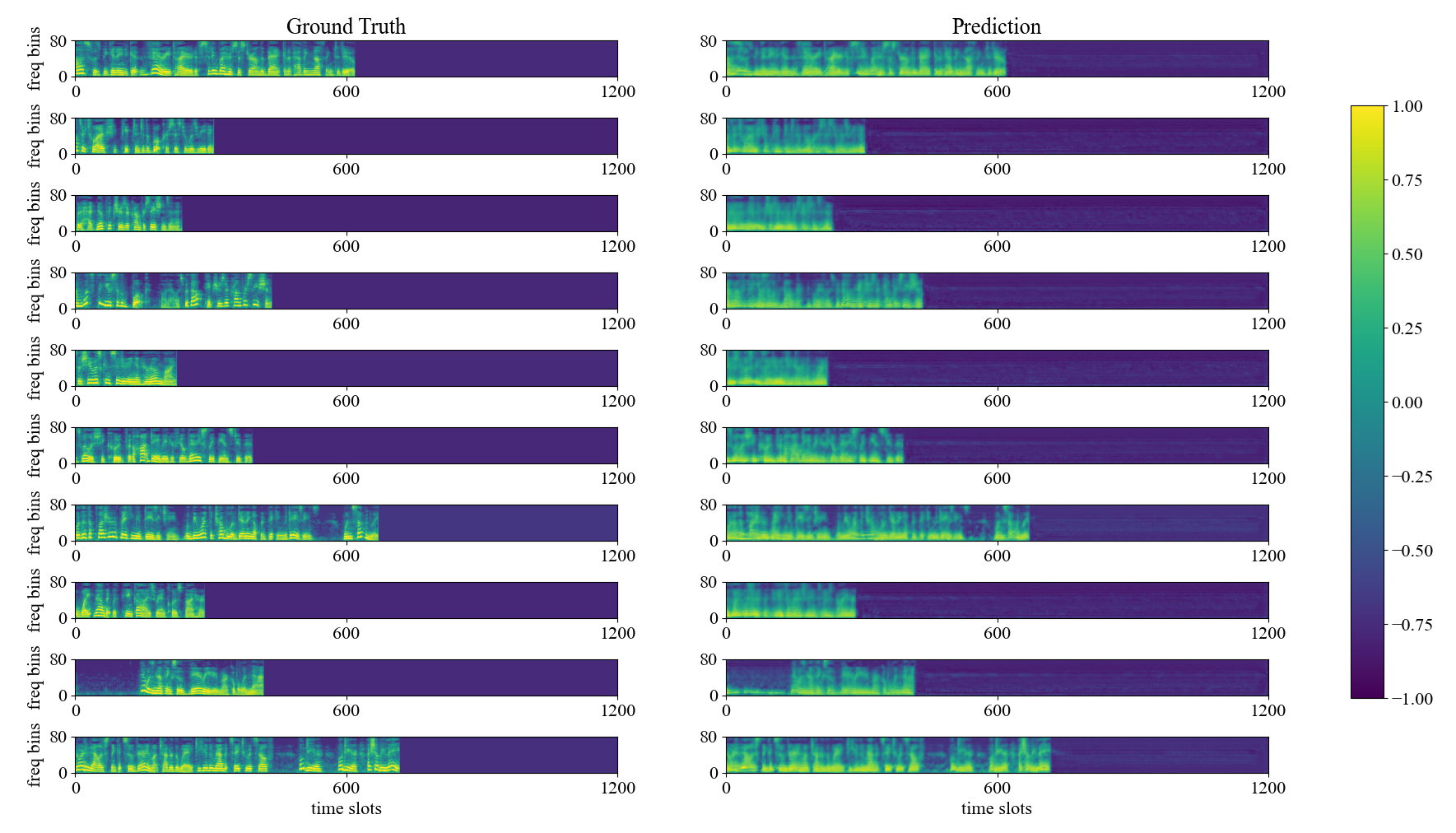} % Reduce the figure size so that it is slightly narrower than the column.
\caption{Predicted Mel spectrograms on the \textit{Brennan} dataset (Left: Ground truth Mel spectrogram; Right: Reconstructed Mel spectrogram).}
\label{fig:brennan_Mel}
\end{figure}

\subsubsection{The Reconstructed Mel Spectrograms}
Figure~\ref{fig:brennan_Mel} shows samples of Mel spectrograms reconstructed from brain signals in the \textit{Brennan} dataset. The corresponding results for the \textit{GWilliams} dataset are provided in Appendix~\ref{sec:samples}.
These samples demonstrate that BrainECHO can produce Mel spectrograms that are largely consistent with the ground truth. Notably, the model effectively restores fine details and accurately predicts the intervals and silent segments in the spectrograms. These results highlight the model's expressive and predictive capabilities, as it can extract Mel spectrograms from brain signal segments exceeding 20 seconds--a feat not achieved by previous methods.

% \subsubsection{Datasets Splitting Strategies}

% \begin{table}[t]
% \small
% \centering
% \resizebox{0.48\textwidth}{!}{
% \begin{tabular}{llcccc}
% \toprule
% \multicolumn{2}{c}{} & \multicolumn{4}{c}{\textbf{BLEU-N (\%) $\uparrow$}} \\
%  \cmidrule{3-6} 
% \textbf{Dataset} & \textbf{Split} & N=1 & N=2 & N=3 & N=4 \\
% \midrule
% \textit{Brennan} & Subject & \textbf{89.78} & \textbf{89.06} & \textbf{88.74} & \textbf{88.55} \\
%  & Sentence & 89.24 & 88.52 & 88.18 & 88.01 \\
% \midrule
% \textit{GWilliams} & RS & 73.35 & 72.66 & 72.46 & 72.42 \\
%  & Session & \textbf{75.24} & \textbf{74.57} & \textbf{74.34} & \textbf{74.27} \\
%  & Subject & 75.05 & 74.38 & 74.18 & 74.14 \\
%  & Sentence & 73.58 & 72.99 & 72.82 & 72.79 \\ 
% \bottomrule
% \end{tabular}
% }
% \centering
% \caption{\label{tab:different_split} Comparison of decoding performance on different datasets and splits. RS denotes random shuffling.}
% \end{table}

% The decoding metrics of BrainECHO across different datasets and splitting strategies are shown in Table~\ref{tab:different_split}. The model demonstrates optimal performance on the \textit{Brennan} and \textit{GWilliams} dataset when split by sentences and sessions, respectively.

\subsection{Ablation and Hyperparameter Study}

\subsubsection{Ablation Study on Three-Stage Training}

\begin{table}[t]
\small
\centering
\resizebox{0.48\textwidth}{!}{
\begin{tabular}{ccccccc}
\toprule
\multicolumn{3}{c}{\textbf{Training Stage}} & \multicolumn{4}{c}{\textbf{BLEU-N (\%) $\uparrow$}} \\
\cmidrule(r){1-3} \cmidrule(l){4-7} 
Au & Al & F & N=1 & N=2 & N=3 & N=4 \\
\midrule
\Checkmark & \Checkmark & \Checkmark & \textbf{89.78} & \textbf{89.06} & \textbf{88.74} & \textbf{88.55} \\
\XSolidBrush & \Checkmark & \Checkmark & 87.13 & 86.29 & 85.92 & 85.74 \\
\XSolidBrush & \XSolidBrush & \Checkmark & 87.63 & 86.87 & 86.54 & 86.38 \\
\Checkmark & \Checkmark & \XSolidBrush & 39.64 & 34.49 & 31.07 & 28.32 \\
\bottomrule
\end{tabular}
}
\centering
\caption{\label{tab:training_stages} Ablation study of training stages on the \textit{Brennan} dataset. The stages labeled Au, Al, and F correspond to Mel autoencoding, brain-audio alignment, and Whisper fine-tuning, respectively.}
\end{table}

To verify the effectiveness of our proposed three-stage training, we incrementally remove each stage and observe the corresponding changes in performance. As presented in Table~\ref{tab:training_stages}, when the autoencoding stage is removed, BLEU-4 drops to 85.74 (-3.17\%). 
Note that in this case, the alignment loss between brain signals and Mel spectrograms in the latent space is removed, while the commitment loss and the reconstruction loss of Mel spectrograms are retained.
Further removal of the brain-audio alignment stage means eliminating the reconstruction loss as well. At this point, the model is trained end-to-end. This leads to an abnormal increase in BLEU, highlighting the challenge of directly constructing a representation space from the brain signals to the Mel spectrogram. However, by pre-warming a discrete representation space, the reconstruction quality and stability are enhanced. 
% In the above two cases, due to the removal of autoencoding in the first stage, the quantizer and audio decoder are randomly initialized and trainable.
% Notably, even without fine-tuning, BrainECHO achieves impressive performance based solely on the predicted Mel spectrogram, suggesting that it is feasible to extract semantically rich audio features from neural signals directly.
% \textcolor{red}{If there is no fine-tuning, i.e., the predicted Mel spectrograms are fed into Whisper directly without fine-tuning in the final stage, the performance is less than ideal. This means that the brain-audio alignment is not perfect, and the audio recognition results from Whisper may deviate from the ground truth. Therefore, the fine-tuning stage is essential to bridge these gaps.}
In the above two cases, the quantizer and audio decoder are randomly initialized and trainable due to the removal of the autoencoding stage. Without fine-tuning in the final stage—i.e., feeding the predicted Mel spectrograms directly into Whisper—the performance is suboptimal. This indicates that the brain-audio alignment is imperfect, and Whisper's recognition results may deviate from the ground truth. Thus, the fine-tuning stage is crucial to bridge these gaps and improve overall performance.

\begin{table}[t]
\small
\centering
\resizebox{0.48\textwidth}{!}{
\begin{tabular}{llcccc}
\toprule
\multicolumn{2}{c}{} & \multicolumn{4}{c}{\textbf{BLEU-N (\%) $\uparrow$}} \\
 \cmidrule{3-6} 
\textbf{Split} & \textbf{Autoencode} & N=1 & N=2 & N=3 & N=4 \\
\midrule
Subject & Separate & 89.78 & 89.06 & \textbf{88.74} & \textbf{88.55} \\
 & Joint & \textbf{89.79} & \textbf{89.08} & 88.73 & \textbf{88.55} \\
\midrule
Sentence & Separate & 89.24 & 88.52 & 88.18 & 88.01 \\
 & Joint & \textbf{89.91} & \textbf{89.22} & \textbf{88.88} & \textbf{88.69} \\
\bottomrule
\end{tabular}
}
\centering
\caption{\label{tab:autoencode} Comparison of decoding performance using separate and joint autoencoding (Separate: Autoencoding trained individually on \textit{Brennan} and \textit{GWilliam} datasets; Joint: Autoencoding trained on the combined \textit{Brennan} and \textit{GWilliam} datasets).
% By default, we employ separate autoencoding.
}
\end{table}

\begin{table}[t]
\small
\centering
\resizebox{0.45\textwidth}{!}{
\begin{tabular}{lcccc}
\toprule
\multicolumn{1}{c}{} & \multicolumn{4}{c}{\textbf{BLEU-N (\%) $\uparrow$}} \\
 \cmidrule{2-5} 
 & N=1 & N=2 & N=3 & N=4 \\
\midrule
w/ quantizer & \textbf{89.78} & \textbf{89.06} & \textbf{88.74} & \textbf{88.55} \\
w/o quantizer & 86.46 & 85.57 & 85.15 & 84.90 \\
\bottomrule
\end{tabular}    
}
\caption{\label{tab:quantizer} Ablation study of quantizer.}
\end{table}

\begin{figure}[t]
\centering
\includegraphics[width=0.48\textwidth]{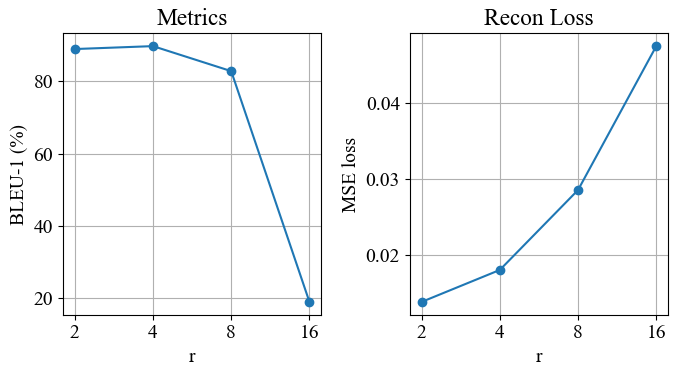} % Reduce the figure size so that it is slightly narrower than the column.
\caption{Changes of BLEU-1 and Mel spectrogram reconstruction loss with different downsampling ratios.}
\label{fig:ratio}
\end{figure}

\subsubsection{Impact of Data Input Strategy in the Autoencoding Stage}

% We compare the decoding performance when the autoencoding task (stage 1) is applied separately to the Mel spectrograms from individual datasets versus both datasets combined. The results, presented in Table~\ref{tab:autoencode}, indicate that joint autoencoding results in either stable or slightly improved metrics (except for BLEU-3) compared to separate autoencoding when splitting \textit{Brennan} by subject. Additionally, all metrics improve when splitting by sentence. This suggests that incorporating Mel spectrograms from other datasets during autoencoding enhances the model's ability to extract richer discrete speech representations, thereby enhancing its generalizability.

This experiment allows us to analyze whether joint training in the autoencoding stage (Stage 1) enhances the model's ability to learn richer and more generalized representations, thereby improving downstream performance in the following stages. Specifically, we compare two approaches: (1) separate autoencoding, where the model is trained individually on the Mel spectrograms from the \textit{Brennan} and \textit{GWilliam} datasets, and (2) joint autoencoding, where the Mel spectrograms from both datasets are combined for training. Following Stage 1, the model proceeds to Stage 2 and Stage 3, which are performed separately on the training sets of \textit{Brennan} and \textit{GWilliam} to evaluate the generalizability and dataset-specific performance. 
The results of the \textit{Brennan} dataset presented in Table~\ref{tab:autoencode} show that, overall, joint autoencoding leads to either stable or slightly improved metrics. However, the improvement is marginal, suggesting that the pre-training datasets need to exhibit high correspondence with the downstream EEG/MEG signals to significantly benefit the decoding framework.

\subsubsection{Hyperparameter Analysis of Audio Encoder Module}
To assess the impact of the downsampling ratio $r$, we evaluate BrainECHO's performance at $r$ values of 2, 4, 8, and 16, while holding other hyperparameters constant. Assuming each pixel in the spectrogram is represented by 8 bits, the corresponding reductions in bit usage are approximately 2.9, 11.6, 46.5, and 186.1, respectively. As illustrated in Figure~\ref{fig:ratio}, increasing $r$ exacerbates information loss, making accurate reconstruction of Mel spectrograms for sentence decoding more challenging. Interestingly, the decoding performance at $r$ = 2 is not as strong as at $r$ = 4, indicating that while a larger feature map enhances reconstruction quality, it may also introduce translation-irrelevant information, thereby complicating the fine-tuning of Whisper. Therefore, selecting a moderate $r$ is essential to optimize latent representation capacity.

\subsubsection{Role of the Discrete Encoding Module}
\label{sec:quantizer}
As shown in Table~\ref{tab:quantizer}, removing the quantizer, i.e., using continuous representation instead of discrete representation, results in a performance decline across all metrics compared to the version with a quantizer. This indicates that discretization can enhance the model's generalization ability by reducing session-specific noise and facilitating the learning of subject-invariant features.

\section{Conclusion}
This paper introduces a novel three-stage brain-to-text framework, BrainECHO, that addresses the shortcomings of prior methods. These methods relied on teacher forcing and failed to compare model performance against pure noise inputs. BrainECHO bridges the latent spaces of text and corresponding aurally evoked brain signals through vector-quantized spectrogram reconstruction and fine-tuned use of the Whisper model. It achieves SOTA performance on public EEG and MEG datasets across various experimental settings.
% By extracting deep semantic information from brain signals, BrainECHO provides valuable insights for future research in the brain-to-text decoding paradigm in the BCI field.
By extracting deep semantic information from brain signals, BrainECHO provides valuable insights for future research in the brain-to-text decoding paradigm in the BCI field.

\section*{Limitations}
The limitations of our proposed work are summarized as follows:
\subsection*{Dataset Limitations}
% Although our method has produced promising results, it is currently suitable only for datasets of audio-evoked neural signals because of the brain-audio feature alignment. Future work can address the limitation by collecting datasets with richer corpora, devising appropriate data augmentation methods, and implementing new modality alignment frameworks.
While our framework successfully achieves sentence-level decoding constrained by a predefined sentence set, it is still far from higher levels of speech/text decoding — specifically, decoding sentences based on known words or even phonemes from the training set. We conducted preliminary experiments on word-level and phoneme-level generalization using a retrieval-augmented generation approach. However, all of these methods have yielded unsatisfactory decoding performance, with BLEU-1 scores not exceeding 10 and BLEU-4 scores approaching zero. We believe this is largely influenced by the dataset paradigm and the amount of data. Specifically, subjects only passively listen to long continuous stories, lacking engagement in multiple modalities. Additionally, sentence lengths vary significantly, making segmentation challenging, and the vocabulary covered in the corpus is extremely limited and unblanced. These factors hinder the advancement of open-vocabulary decoding paradigms. We encourage future researchers (including ourselves) to collect larger-scale, multimodal datasets with a well-structured stimulus presentation. Such datasets would lay a solid foundation for designing more generalizable and robust decoding frameworks.
\subsection*{Experiment Limitations}
In our experimental setting, all data are strictly segmented on a sentence-by-sentence basis before being fed into the model, which may not align with real-world decoding scenarios, due to the potential unknown length of the signals to be translated. 
Moreover, according to the results reported by NeuSpeech~\citep{yang2024decode}, sentence-level decoding may face overfitting issues, as neural signals of different lengths need to be padded to the same length before fed into the model. However, under the condition that there is a correlation between signal length and sentence length, our approach may help the model decode by implicitly injecting the length information of the signal. 
Moreover, as reported by NeuSpeech~\citep{yang2024decode}, sentence-level decoding might encounter overfitting problems. The reason is that neural signals of varying lengths should be padded to a consistent length before being fed into the model. When a correlation exists between signal length and sentence length, it is possible that our proposed approach inadvertently facilitates the model's decoding by implicitly integrating the length information of the signal. MAD~\citep{yang2024mad} and NeuGPT~\citep{yang2024neugptunifiedmultimodalneural} showed an unsatisfactory result with a uniform signal length, suggesting that the current task of generating open-vocabulary text based solely on the neural signal pattern remains extremely challenging. Our forthcoming research efforts will focus on leveraging LLMs and more efficient alignment strategies to diminish the dependence on length information.

\section*{Ethical Statement}
This study uses publicly available datasets and does not involve the collection of any brain activity data from human subjects. Therefore, our research does not have any adverse impact on human society.

\section*{Acknowledgements}
This work is supported by the National Natural Science Foundation of China (Grant No. 62306089), the Shenzhen Science and Technology Program (Grant Nos. RCBS20231211090800003 and ZDSYS20230626091203008) and the Shenzhen-Hong Kong Institute of Brain Science-Shenzhen Fundamental Research Institutions (2023SHIBS0003).

\bibliography{custom}

\begin{thebibliography}{34}
\providecommand{\natexlab}[1]{#1}

\bibitem[{Amrani et~al.(2024)Amrani, Micucci, and Napoletano}]{amrani2024deep}
Hamza Amrani, Daniela Micucci, and Paolo Napoletano. 2024.
\newblock Deep representation learning for open vocabulary electroencephalography-to-text decoding.
\newblock \emph{IEEE Journal of Biomedical and Health Informatics}.

\bibitem[{Baevski et~al.(2020)Baevski, Zhou, Mohamed, and Auli}]{baevski2020wav2vec}
Alexei Baevski, Yuhao Zhou, Abdelrahman Mohamed, and Michael Auli. 2020.
\newblock wav2vec 2.0: A framework for self-supervised learning of speech representations.
\newblock \emph{Advances in neural information processing systems}, 33:12449--12460.

\bibitem[{Bain et~al.(2023)Bain, Huh, Han, and Zisserman}]{bain2023whisperx}
Max Bain, Jaesung Huh, Tengda Han, and Andrew Zisserman. 2023.
\newblock Whisperx: Time-accurate speech transcription of long-form audio.
\newblock \emph{arXiv preprint arXiv:2303.00747}.

\bibitem[{Brennan and Hale(2019)}]{brennan2019hierarchical}
Jonathan~R Brennan and John~T Hale. 2019.
\newblock Hierarchical structure guides rapid linguistic predictions during naturalistic listening.
\newblock \emph{PloS one}, 14(1):e0207741.

\bibitem[{Chen et~al.(2024{\natexlab{a}})Chen, Du, Liu, Wang, and He}]{chen2024open}
Xiaoyu Chen, Changde Du, Che Liu, Yizhe Wang, and Huiguang He. 2024{\natexlab{a}}.
\newblock Open-vocabulary auditory neural decoding using fmri-prompted llm.
\newblock \emph{arXiv preprint arXiv:2405.07840}.

\bibitem[{Chen et~al.(2024{\natexlab{b}})Chen, Wang, Khalilian-Gourtani, Yu, Dugan, Friedman, Doyle, Devinsky, Wang, and Flinker}]{chen2024neural}
Xupeng Chen, Ran Wang, Amirhossein Khalilian-Gourtani, Leyao Yu, Patricia Dugan, Daniel Friedman, Werner Doyle, Orrin Devinsky, Yao Wang, and Adeen Flinker. 2024{\natexlab{b}}.
\newblock A neural speech decoding framework leveraging deep learning and speech synthesis.
\newblock \emph{Nature Machine Intelligence}, pages 1--14.

\bibitem[{D{\'e}fossez et~al.(2023)D{\'e}fossez, Caucheteux, Rapin, Kabeli, and King}]{defossez2023decoding}
Alexandre D{\'e}fossez, Charlotte Caucheteux, J{\'e}r{\'e}my Rapin, Ori Kabeli, and Jean-R{\'e}mi King. 2023.
\newblock Decoding speech perception from non-invasive brain recordings.
\newblock \emph{Nature Machine Intelligence}, 5(10):1097--1107.

\bibitem[{Duan et~al.(2024)Duan, Chau, Wang, Wang, and Lin}]{duan2024dewave}
Yiqun Duan, Charles Chau, Zhen Wang, Yu-Kai Wang, and Chin-teng Lin. 2024.
\newblock Dewave: Discrete encoding of eeg waves for eeg to text translation.
\newblock \emph{Advances in Neural Information Processing Systems}, 36.

\bibitem[{Feng et~al.(2023)Feng, Feng, Qin, and Liu}]{feng2023aligning}
Xiachong Feng, Xiaocheng Feng, Bing Qin, and Ting Liu. 2023.
\newblock Aligning semantic in brain and language: A curriculum contrastive method for electroencephalography-to-text generation.
\newblock \emph{IEEE Transactions on Neural Systems and Rehabilitation Engineering}.

\bibitem[{Ghazaryan et~al.(2023)Ghazaryan, van Vliet, Saranp{\"a}{\"a}, Lammi, Lindh-Knuutila, Hult{\'e}n, Kivisaari, and Salmelin}]{ghazaryan2023trials}
Gayane Ghazaryan, Marijn van Vliet, Aino Saranp{\"a}{\"a}, Lotta Lammi, Tiina Lindh-Knuutila, Annika Hult{\'e}n, Sasa Kivisaari, and Riitta Salmelin. 2023.
\newblock Trials and tribulations when attempting to decode semantic representations from meg responses to written text.
\newblock \emph{Language, Cognition and Neuroscience}, pages 1--12.

\bibitem[{Gwilliams et~al.(2023)Gwilliams, Flick, Marantz, Pylkk{\"a}nen, Poeppel, and King}]{gwilliams2023introducing}
Laura Gwilliams, Graham Flick, Alec Marantz, Liina Pylkk{\"a}nen, David Poeppel, and Jean-R{\'e}mi King. 2023.
\newblock Introducing meg-masc a high-quality magneto-encephalography dataset for evaluating natural speech processing.
\newblock \emph{Scientific data}, 10(1):862.

\bibitem[{Jo et~al.(2024)Jo, Yang, Han, Duan, Xiong, and Lee}]{jo2024eeg}
Hyejeong Jo, Yiqian Yang, Juhyeok Han, Yiqun Duan, Hui Xiong, and Won~Hee Lee. 2024.
\newblock Are eeg-to-text models working?
\newblock \emph{arXiv preprint arXiv:2405.06459}.

\bibitem[{Kong et~al.(2021)Kong, Cao, Liu, Choi, and Wang}]{kong2021decoupling}
Qiuqiang Kong, Yin Cao, Haohe Liu, Keunwoo Choi, and Yuxuan Wang. 2021.
\newblock Decoupling magnitude and phase estimation with deep resunet for music source separation.
\newblock In \emph{22nd International Conference on Music Information Retrieval, ISMIR 2021}, pages 342--349. International Society for Music Information Retrieval.

\bibitem[{Lewis et~al.(2020)Lewis, Liu, Goyal, Ghazvininejad, Mohamed, Levy, Stoyanov, and Zettlemoyer}]{lewis2020bart}
Mike Lewis, Yinhan Liu, Naman Goyal, Marjan Ghazvininejad, Abdelrahman Mohamed, Omer Levy, Veselin Stoyanov, and Luke Zettlemoyer. 2020.
\newblock Bart: Denoising sequence-to-sequence pre-training for natural language generation, translation, and comprehension.
\newblock In \emph{Proceedings of the 58th Annual Meeting of the Association for Computational Linguistics}, pages 7871--7880.

\bibitem[{Li et~al.(2023)Li, Jia, and Chiu}]{li2023textless}
Xinjian Li, Ye~Jia, and Chung-Cheng Chiu. 2023.
\newblock Textless direct speech-to-speech translation with discrete speech representation.
\newblock In \emph{ICASSP 2023-2023 IEEE International Conference on Acoustics, Speech and Signal Processing (ICASSP)}, pages 1--5. IEEE.

\bibitem[{Lin(2004)}]{lin2004rouge}
Chin-Yew Lin. 2004.
\newblock Rouge: A package for automatic evaluation of summaries.
\newblock In \emph{Text summarization branches out}, pages 74--81.

\bibitem[{Metzger et~al.(2023)Metzger, Littlejohn, Silva, Moses, Seaton, Wang, Dougherty, Liu, Wu, Berger et~al.}]{metzger2023high}
Sean~L Metzger, Kaylo~T Littlejohn, Alexander~B Silva, David~A Moses, Margaret~P Seaton, Ran Wang, Maximilian~E Dougherty, Jessie~R Liu, Peter Wu, Michael~A Berger, et~al. 2023.
\newblock A high-performance neuroprosthesis for speech decoding and avatar control.
\newblock \emph{Nature}, 620(7976):1037--1046.

\bibitem[{Papineni et~al.(2002)Papineni, Roukos, Ward, and Zhu}]{papineni2002bleu}
Kishore Papineni, Salim Roukos, Todd Ward, and Wei-Jing Zhu. 2002.
\newblock Bleu: a method for automatic evaluation of machine translation.
\newblock In \emph{Proceedings of the 40th annual meeting of the Association for Computational Linguistics}, pages 311--318.

\bibitem[{Radford et~al.(2023)Radford, Kim, Xu, Brockman, McLeavey, and Sutskever}]{radford2023robust}
Alec Radford, Jong~Wook Kim, Tao Xu, Greg Brockman, Christine McLeavey, and Ilya Sutskever. 2023.
\newblock Robust speech recognition via large-scale weak supervision.
\newblock In \emph{International conference on machine learning}, pages 28492--28518. PMLR.

\bibitem[{Razavi et~al.(2019)Razavi, van~den Oord, and Vinyals}]{razavi2019generating}
Ali Razavi, A{\"a}ron van~den Oord, and Oriol Vinyals. 2019.
\newblock Generating diverse high-fidelity images with vq-vae-2.
\newblock In \emph{Proceedings of the 33rd International Conference on Neural Information Processing Systems}, pages 14866--14876.

\bibitem[{Sadok et~al.(2023)Sadok, Leglaive, and S{\'e}guier}]{sadok2023vector}
Samir Sadok, Simon Leglaive, and Renaud S{\'e}guier. 2023.
\newblock A vector quantized masked autoencoder for speech emotion recognition.
\newblock In \emph{2023 IEEE International conference on acoustics, speech, and signal processing workshops (ICASSPW)}, pages 1--5. IEEE.

\bibitem[{Shirakawa et~al.(2024)Shirakawa, Nagano, Tanaka, Aoki, Majima, Muraki, and Kamitani}]{shirakawa2024spurious}
Ken Shirakawa, Yoshihiro Nagano, Misato Tanaka, Shuntaro~C Aoki, Kei Majima, Yusuke Muraki, and Yukiyasu Kamitani. 2024.
\newblock Spurious reconstruction from brain activity: The thin line between reconstruction, classification, and hallucination.
\newblock \emph{Journal of Vision}, 24(10):321--321.

\bibitem[{Song et~al.(2022)Song, Zheng, Liu, and Gao}]{song2022eeg}
Yonghao Song, Qingqing Zheng, Bingchuan Liu, and Xiaorong Gao. 2022.
\newblock Eeg conformer: Convolutional transformer for eeg decoding and visualization.
\newblock \emph{IEEE Transactions on Neural Systems and Rehabilitation Engineering}, 31:710--719.

\bibitem[{Tang et~al.(2023)Tang, LeBel, Jain, and Huth}]{tang2023semantic}
Jerry Tang, Amanda LeBel, Shailee Jain, and Alexander~G Huth. 2023.
\newblock Semantic reconstruction of continuous language from non-invasive brain recordings.
\newblock \emph{Nature Neuroscience}, 26(5):858--866.

\bibitem[{Van Den~Oord et~al.(2017)Van Den~Oord, Vinyals et~al.}]{van2017neural}
Aaron Van Den~Oord, Oriol Vinyals, et~al. 2017.
\newblock Neural discrete representation learning.
\newblock \emph{Advances in neural information processing systems}, 30.

\bibitem[{Wang and Ji(2022)}]{wang2022open}
Zhenhailong Wang and Heng Ji. 2022.
\newblock Open vocabulary electroencephalography-to-text decoding and zero-shot sentiment classification.
\newblock In \emph{Proceedings of the AAAI Conference on Artificial Intelligence}, pages 5350--5358.

\bibitem[{Willett et~al.(2023)Willett, Kunz, Fan, Avansino, Wilson, Choi, Kamdar, Glasser, Hochberg, Druckmann et~al.}]{willett2023high}
Francis~R Willett, Erin~M Kunz, Chaofei Fan, Donald~T Avansino, Guy~H Wilson, Eun~Young Choi, Foram Kamdar, Matthew~F Glasser, Leigh~R Hochberg, Shaul Druckmann, et~al. 2023.
\newblock A high-performance speech neuroprosthesis.
\newblock \emph{Nature}, 620(7976):1031--1036.

\bibitem[{Xi et~al.(2023)Xi, Zhao, Wang, Liu, Qin, and Liu}]{xi2023unicorn}
Nuwa Xi, Sendong Zhao, Haochun Wang, Chi Liu, Bing Qin, and Ting Liu. 2023.
\newblock Unicorn: Unified cognitive signal reconstruction bridging cognitive signals and human language.
\newblock \emph{arXiv preprint arXiv:2307.05355}.

\bibitem[{Yang et~al.(2023)Yang, Yu, Wang, Wang, Weng, Zou, and Yu}]{yang2023diffsound}
Dongchao Yang, Jianwei Yu, Helin Wang, Wen Wang, Chao Weng, Yuexian Zou, and Dong Yu. 2023.
\newblock Diffsound: Discrete diffusion model for text-to-sound generation.
\newblock \emph{IEEE/ACM Transactions on Audio, Speech, and Language Processing}, 31:1720--1733.

\bibitem[{Yang et~al.(2024{\natexlab{a}})Yang, Duan, Jo, Zhang, Xu, Jones, Hu, teng Lin, and Xiong}]{yang2024neugptunifiedmultimodalneural}
Yiqian Yang, Yiqun Duan, Hyejeong Jo, Qiang Zhang, Renjing Xu, Oiwi~Parker Jones, Xuming Hu, Chin teng Lin, and Hui Xiong. 2024{\natexlab{a}}.
\newblock \href {https://arxiv.org/abs/2410.20916} {Neugpt: Unified multi-modal neural gpt}.
\newblock \emph{Preprint}, arXiv:2410.20916.

\bibitem[{Yang et~al.(2024{\natexlab{b}})Yang, Duan, Zhang, Xu, and Xiong}]{yang2024decode}
Yiqian Yang, Yiqun Duan, Qiang Zhang, Renjing Xu, and Hui Xiong. 2024{\natexlab{b}}.
\newblock Decode neural signal as speech.
\newblock \emph{arXiv preprint arXiv:2403.01748}.

\bibitem[{Yang et~al.(2024{\natexlab{c}})Yang, Jo, Duan, Zhang, Zhou, Lee, Xu, and Xiong}]{yang2024mad}
Yiqian Yang, Hyejeong Jo, Yiqun Duan, Qiang Zhang, Jinni Zhou, Won~Hee Lee, Renjing Xu, and Hui Xiong. 2024{\natexlab{c}}.
\newblock Mad: Multi-alignment meg-to-text decoding.
\newblock \emph{arXiv preprint arXiv:2406.01512}.

\bibitem[{Zhang et~al.(2023)Zhang, Chen, Bukharin, Karampatziakis, He, Cheng, Chen, and Zhao}]{zhang2023adalora}
Qingru Zhang, Minshuo Chen, Alexander Bukharin, Nikos Karampatziakis, Pengcheng He, Yu~Cheng, Weizhu Chen, and Tuo Zhao. 2023.
\newblock Adalora: Adaptive budget allocation for parameter-efficient fine-tuning.
\newblock \emph{arXiv preprint arXiv:2303.10512}.

\bibitem[{Zhao et~al.(2024)Zhao, Sun, Wang, Ye, Xhz, and Zong}]{zhao2024mapguide}
Xinpei Zhao, Jingyuan Sun, Shaonan Wang, Jing Ye, Xhz Xhz, and Chengqing Zong. 2024.
\newblock Mapguide: A simple yet effective method to reconstruct continuous language from brain activities.
\newblock In \emph{Proceedings of the 2024 Conference of the North American Chapter of the Association for Computational Linguistics: Human Language Technologies (Volume 1: Long Papers)}, pages 3822--3832.

\end{thebibliography}

\appendix

\section{Conformer}
\label{sec:conformer}
Conformer utilizes a Convolution-Transformer architecture to capture both local and global features. The one-dimensional temporal and spatial convolution layers in TS Conv capture the local information of neural signals, while the self-attention modules in the Transformer blocks extract the global dependencies of these local time features. The detailed structure of Conformer is provided in Table~\ref{tab:conformer}.

\begin{table*}[t]
\small
\resizebox{1.0\textwidth}{!}{
\centering
\begin{tabular}{c|cccccc}
\toprule
\textbf{Layer Type} & \textbf{Out Channels} & \textbf{Filter Size} & \textbf{Stride} & \textbf{Padding} & \textbf{Input} & \textbf{Output} \\
\midrule 
Conv2D & 64 & (1, 5) & (1, 2) & 2 & $1\times C\times T_\varepsilon$ & $64\times C\times \frac{T_\varepsilon}{2}$ \\
\rule{0pt}{10pt}
BatchNorm2D + ELU & - & - & - & - & $64\times C\times \frac{T_\varepsilon}{2}$ & $64\times C\times \frac{T_\varepsilon}{2}$ \\
\midrule
Conv2D & 128 & (1, 3) & (1, 2) & 1 & $64\times C\times \frac{T_\varepsilon}{2}$ & $128\times C\times \frac{T_\varepsilon}{4}$ \\
\rule{0pt}{10pt}
BatchNorm2D + ELU & - & - & - & - & $128\times C\times \frac{T_\varepsilon}{4}$ & $128\times C\times \frac{T_\varepsilon}{4}$ \\
\midrule
Conv2D & 256 & ($C$, 1) & 1 & 0 & $128\times C\times \frac{T_\varepsilon}{4}$ & $256\times C\times \frac{T_\varepsilon}{4}$ \\
\rule{0pt}{10pt}
BatchNorm2D + ELU & - & - & - & - & $256\times C\times \frac{T_\varepsilon}{4}$ & $256\times 1\times \frac{T_\varepsilon}{4}$ \\
\midrule
Rearrange & - & - & - & - & $256\times 1\times \frac{T_\varepsilon}{4}$ & $\frac{T_\varepsilon}{4}\times 256$ \\
\bottomrule
\end{tabular}  
}
\centering
\caption{\label{tab:conformer} The structure of TS Conv. $C$ and $T_\varepsilon$ denote the number of EEG/MEG channels and timestamps.}
\end{table*}

\begin{table*}[!t]
    \small
    \resizebox{1.0\textwidth}{!}{
\begin{tabular}{llp{.65\textwidth}l}
\toprule
\textbf{Dataset} & \textbf{Split} & \textbf{Details} & \textbf{Result} \\ \cmidrule{1-4}
\textit{Brennan} & Sentence & For each participant, sentence-EEG/MEG pairs corresponding to random selected 10\% of unique sentences are allocated to the test set, then the remaining sentence-EEG/MEG pairs are shuffled and split into train:valid 8:1. Note that the test set for each subject may contain different sentences and the training set may cover all possible sentences. & 3696:462:462 \\ \cmidrule{2-4}
 & Subject & 3 participants (about 10\% of the total number of subjects) are selected at random for the test set, 3 for the validation set, and the remaining 27 for the training set. & 3780:420:420 \\ \cmidrule{1-4}
\textit{GWilliams} & RS & All data is random shuffled and divided into train:valid:test 8:1:1. & 23339:2917:2918 \\ \cmidrule{2-4}
 & Session & Random shuffled data of session 0 is divided into train:valid 8:1 and data of session 1 is held out as test set. & 13129:2976:13069 \\ \cmidrule{2-4}
 & Sentence & It is the same as \textit{Brennan} above. & 23305:2914:2955 \\ \cmidrule{2-4}
 & Subject & 2 participants (about 10\% of the total number of subjects) are selected at random for the test set, 2 for the validation set, and the remaining 23 for the training set. & 24137:2469:2568 \\ \bottomrule
\end{tabular}
    }
\centering
\caption{Details of different dataset split settings. RS denotes random shuffling. \label{tab:dataset_split}}
\end{table*}

\section{Necessity of the Three-Stage Paradigm}
\label{sec:necessity}
Although it is possible to integrate Stage 2 and Stage 3, or even all stages, into a single stage for end-to-end training, we chose to adopt a three-stage training paradigm for several reasons.

First, our three-stage design achieves decoupled representation learning, and each stage serves a distinct and crucial purpose. Stage 1 focuses on optimizing the quantizer, audio encoder and decoder. Stage 2 is dedicated to aligning brain signals with Mel spectrograms. By doing so, we can better capture the complex relationship between the neural activity and the corresponding audio features, which is a key step in bridging the gap between brain signals and text. Finally, in Stage 3, we leverage the Mel spectrograms as an intermediate modality. Through fine-tuning Whisper, we are able to align the brain signals with the text modality. By decoupling brain signal representation learning from linguistic knowledge preservation, we mitigate the risk of LLM over-dominance and establish robust EEG/MEG-to-text mapping.

Second, with only a single Nvidia 3090 (24GB) GPU, we found that it would be challenging to support the large-scale parameter training required for the combined stage. One of the advantages of this three-stage design is that each stage only requires the optimization of a specific subset of module parameters. This significantly reduces the computational burden at each step, enabling us to effectively balance the decoding performance and resource utilization. As demonstrated by our experimental results in Section~\ref{sec:results}, this three-stage training approach is highly effective, highlighting its potential for similar research in the field.

\section{Datasets}
\label{sec:datasets}
\subsection{\textit{Brennan}}
% \setlabel{B.1}{sec:brennan}
The \textit{Brennan} dataset~\cite{brennan2019hierarchical} contains raw electroencephalography (EEG) data collected from 49 human subjects. Participants were asked to passively listen to a 12.4-minute audiobook story of chapter one of \textit{Alice’s Advenctures in Wonderland}, while their EEG data was recorded. Participants completed an eight-question multiple choice questionnaire concerning the contents of the story at the end of the experimental session. We retain 33 participants' data who achieved high scores.

Participants were fitted with an elastic cap with 61 actively-amplified electrodes and one ground electrode (actiCap, Brain Products GmbH). Electrodes were distributed equidistantly across the scalp according to the Easycap M10 layout. Conductive gel was inserted into each electrode to reduce impedences to 25 kOhms or below. Data were recorded at 500 Hz between 0.1 and 200 Hz referenced to an electrode placed on the right mastoid (actiCHamp, Brain Products GmbH).

The stimulus chapter originally contains 84 sentences. Since the annotation files only provide word-level annotations, directly concatenating words to form sentences would result in the absence of punctuation marks. Therefore, we use WhisperX~\cite{bain2023whisperx} to segment the audio stimulus into segments of no more than 12 seconds, resulting in 140 sentences.

\subsection{\textit{GWilliams}}
% \setlabel{B.2}{sec:gwilliams}
\textit{GWilliams}~\cite{gwilliams2023introducing}, known as the ``MEG-MASC'' dataset, provides raw magnetoencephalography (MEG) data from 27 English speakers who listened to two hours of naturalistic stories. Each participant performed two identical sessions, involving listening to four fictional stories from the Manually Annotated Sub-Corpus (MASC). The four stories are: `LW1' (861 words, 5 min 20 sec), `Cable Spool Boy' (1948 words, 11 min), `Easy Money' (3541 words, 12 min 10 sec) and `The Black Willow' (4652 words, 25 min 50 sec).

An audio track corresponding to each of these stories was synthesized using Mac OS Mojave © version 10.14 text-to-speech. To help decorrelate language features from acoustic representations, both voices and speech rate were varied every 5–20 sentences. Specifically, three distinct synthetic voices: `Ava', `Samantha' and `Allison' are used speaking between 145 and 205 words per minute. Additionally, the silence between sentences are varied between 0 and 1,000ms. Both speech rate and silence duration were sampled from a uniform distribution between the min and max values.

Each story was divided into $\sim$3 min sound files. In between these sounds— approximately every 30 s— a random word list generated from the unique content words (nouns, proper nouns, verbs, adverbs and adjectives) selected from the preceding 5min segment presented in random order were played.

Within each $\sim$1 h recording session, participants were recorded with a 208 axial-gradiometer MEG scanner built by the Kanazawa Institute of Technology (KIT), and sampled at 1,000 Hz, and online band-pass fltered between 0.01 and 200Hz while they listened to four distinct stories through binaural tube earphones (Aero Technologies), at a mean level of 70dB sound pressure level.

To ensure a fair comparison with NeuSpeech~\cite{yang2024decode}, we follow its experimental setup by concatenating words with the same sentence ID into full sentences, based on the annotation files. This process results in 661 sentences.

\begin{table*}[!t]
\small
\resizebox{1.0\textwidth}{!}{
\begin{tabular}{ccccccc}
\toprule
\multicolumn{1}{c}{} & \multicolumn{3}{c}{\textit{Brennan}} & \multicolumn{3}{c}{\textit{GWilliams}} \\
\cmidrule(r){2-4} \cmidrule(l){5-7} 
\textbf{Configuration} & Pretraining & Alignment & Finetuning & Pretraining & Alignment & Finetuning \\
\midrule
Batch Size & 16 & 16 & 16 & 16 & 8 & 16 \\
Max Epoch & 400 & 40 & 40 & 100 & 40 & 40 \\
Max Learning Rate & 2e-4 & 1e-4 & 1e-4 & 2e-4 & 1e-4 & 2e-4 \\
Optimizer & \multicolumn{6}{c}{AdamW, with weight decay = 1e-2, betas = (0.9,0.999)} \\
LR Scheduler & \multicolumn{6}{c}{Cosine Annealing, with T\_max = Max Epoch} \\
Early Stopping Patience & \multicolumn{6}{c}{4} \\
\bottomrule
\end{tabular}
}
\centering
\caption{\label{tab:imp_details} Details of the experimental configuration.}
\end{table*}

\begin{figure*}[t]
\centering
\includegraphics[width=1.0\textwidth]{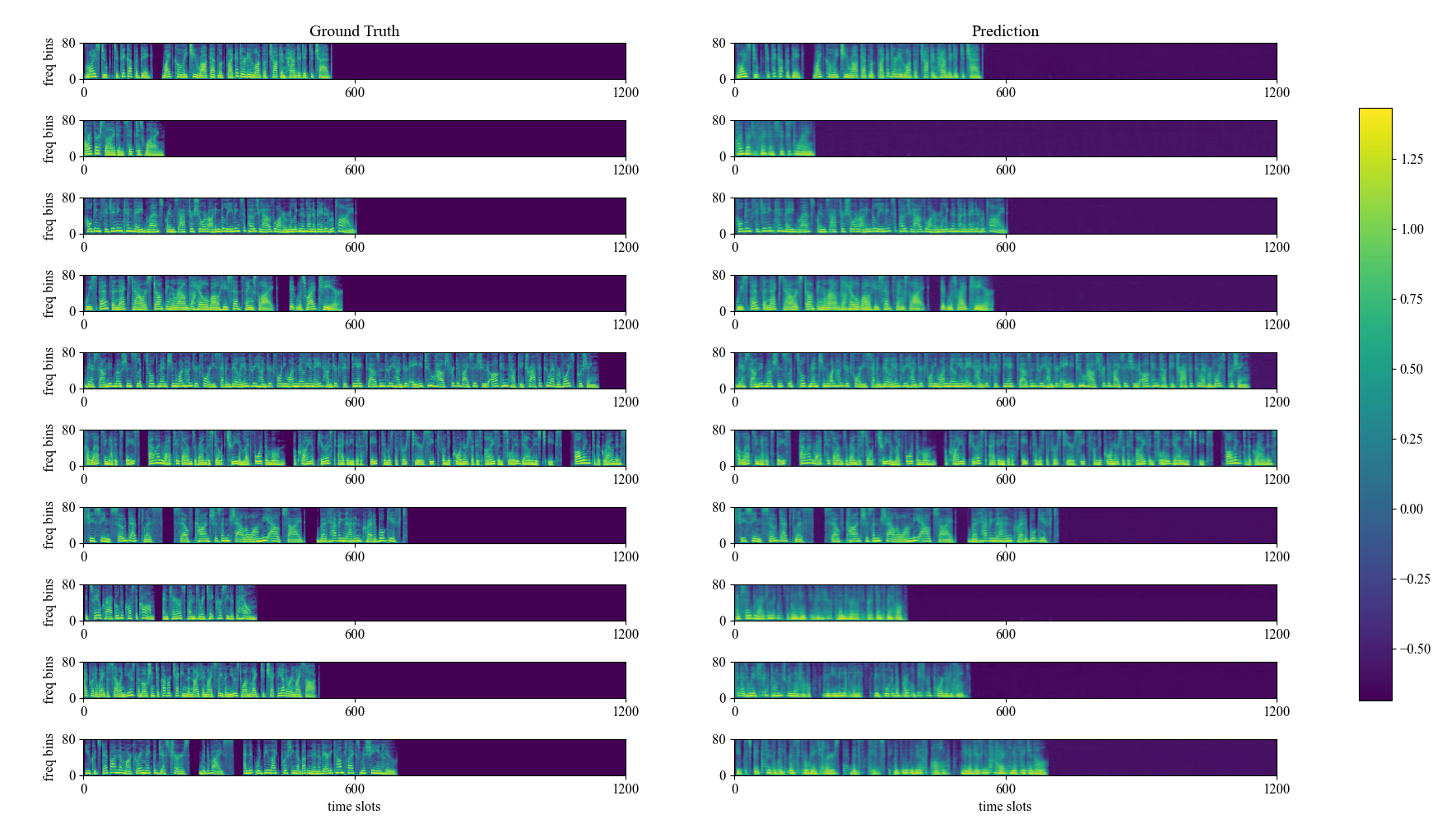}
\caption{Predicted Mel spectrograms on the \textit{GWilliams} dataset.
% For visualization purposes, only the first half of the Mel spectrograms are displayed.
}
\label{fig:gwilliams_Mel}
\end{figure*}

\begin{table*}[!t]
\small 
\centering
\resizebox{1.0\textwidth}{!}{
\begin{tabular}{llp{0.82\textwidth}}              
\toprule
\multicolumn{3}{l}{Generated samples on \textit{Brennan}} 
\\ \cmidrule{1-3}
{\multirow{5}{*}{(1)}} & Ground Truth                & There seemed to be no use in waiting by the little door, so she went back to the table.                                                             \\ 
\multicolumn{1}{c}{}                     & EEG-to-Text         & But they were all locked, and when Alice had been all the way down one side and up the other trying every door, she did not care how she was ever to get out again.                              \\
\multicolumn{1}{c}{}                     & NeuSpeech   & \textbf{There seemed to be no use in waiting by the little door, so she went back to the table.}                                           \\
\multicolumn{1}{c}{}                     & BrainECHO \textit{w/o ft} & \textbf{There seemed to be no use in waiting by the little door, so she went back to the table.}        \\
\multicolumn{1}{c}{}                     & BrainECHO & \textbf{There seemed to be no use in waiting by the little door, so she went back to the table.}        \\ \cmidrule{1-3} 
\multirow{5}{*}{(2)}                      & Ground Truth                & that she'd never before seen a rabbit with either a waistcoat pocket or a watch to take out of it, and burning with curiosity, she ran across the field after it, and fortunately                                                             \\ 
\multicolumn{1}{c}{}                     & EEG-to-Text         & how she longed to get out of that dark hall and wander about among those beds of bright flowers and those cool fountains, but she did not even get her head through the doorway.                              \\
\multicolumn{1}{c}{}                     & NeuSpeech   & But they were all locked, and when Alice had been all the way down one side and up the other trying every door, she walked sadly down the middle, wondering how she was ever to get out again.                                           \\
\multicolumn{1}{c}{}                     & BrainECHO \textit{w/o ft} & But \textbf{she will never} be foreseen around it, \textbf{with either a waistcoat pocket or a watch to take out of it and burn} in \textbf{curiosity. She ran across the field after it} unfortunately.        \\
\multicolumn{1}{c}{}                     & BrainECHO & \textbf{that she'd never before seen a rabbit with either a waistcoat pocket or a watch to take out of it and burning with curiosity, she ran across the field after it, and fortunately}                                                 \\ \bottomrule
\toprule
\multicolumn{3}{l}{Generated samples on \textit{GWilliams}} 
\\ \cmidrule{1-3}
{\multirow{5}{*}{(1)}} & Ground Truth                & I seen him since high school maybe twenty years before and we were never buddies in the first place                                                             \\ 
\multicolumn{1}{c}{}                     & EEG-to-Text         & \underline{It was a long time since I had last seen him} in the flesh                              \\
\multicolumn{1}{c}{}                     & NeuSpeech   & \textbf{I seen him since high school} \underline{when I was young}, at least \textbf{before and we were never buddies in} any place.                                           \\
\multicolumn{1}{c}{}                     & BrainECHO \textit{w/o ft} & I hadn't \textbf{seen him since high school, maybe 20 years before} and you remember when he's \textbf{in the first place}.        \\
\multicolumn{1}{c}{}                     & BrainECHO & \textbf{I seen him since high school maybe twenty years before and we were never buddies in the first place}        \\ \cmidrule{1-3} 
\multirow{5}{*}{(2)}                      & Ground Truth                & My patience was long gone and I was back in the car to warming up when Acres tapped on the window and told me he had found whatever he was looking for                                                             \\ 
\multicolumn{1}{c}{}                     & EEG-to-Text         & \underline{He said} he had no idea how \textbf{long} it would take him to get \textbf{back} home                              \\
\multicolumn{1}{c}{}                     & NeuSpeech   & \textbf{My patience was long gone and I was back in the car}. But when I heard that many of you were \textbf{looking for} whatever it was, but what about this?                                           \\
\multicolumn{1}{c}{}                     & BrainECHO \textit{w/o ft} & \textbf{My patience was long gone}, \textbf{and I was back in the car to warming up when acres tapped on the window and} Tunch \textbf{told me he had found whatever he was looking for}.        \\
\multicolumn{1}{c}{}                     & BrainECHO & \textbf{My patience was long gone and I was back in the car to warming up when Acres tapped on the window and told me he had found whatever he was looking for}                                                 \\ \bottomrule
\end{tabular}
}

\centering
\caption{
Comparison of decoding sentences generated by different methods, where \textbf{bold} and \underline{underline} indicate an exact match and a similar match, respectively, between prediction and ground truth. All methods use the same generation configuration. \textit{w/o ft} means decoding by inputting the predicted Mel spectrogram into Whisper directly without fine-tuning in the final stage. Only examples of NeuSpeech are reported rather than those of MAD because of NeuSpeech's overall superior performance and the similarity of its method to MAD's.
\label{tab:generated_samples} }
\end{table*}

\begin{table*}[!t]
\small
\resizebox{1.0\textwidth}{!}{
\begin{tabular}{llp{14cm}}              
\toprule
{\multirow{2}{*}{(1)}} & Ground Truth                & There were doors all around the hall.                                                             \\ \cmidrule{2-3}
\multicolumn{1}{c}{}                     & Predicted & not much larger than a rat hole.        \\ \cmidrule{1-3} 
\multirow{2}{*}{(2)}                      & Ground Truth                & For you see, as she couldn't answer either question, it didn't much matter which way she put it.                  \\ \cmidrule{2-3}
\multicolumn{1}{c}{}                     & Predicted & \textbf{For you see, as she couldn't answer either question, it didn't much matter which way she put it.}                                        \\ \cmidrule{1-3} 
\multirow{2}{*}{(3)}                      & Ground Truth                & When she thought it over afterwards, it occurred to her that she ought to have wondered at this, but at the time it all seemed quite natural.                  \\ \cmidrule{2-3}
\multicolumn{1}{c}{}                     & Predicted & \textbf{When she thought it over afterwards, it occurred to her that she ought to have wondered at this, but at the time it all seemed quite natural.}                                        \\ \cmidrule{1-3} 
\multirow{2}{*}{(4)}                      & Ground Truth                & I wonder how many miles I've fallen by this time, she said aloud.                  \\ \cmidrule{2-3}
\multicolumn{1}{c}{}                     & Predicted & \textbf{I wonder how many miles I've fallen by this time, she said aloud.}                                        \\ \cmidrule{1-3} 
\multirow{2}{*}{(5)}                      & Ground Truth                & and that if you cut your finger very deeply with a knife, it usually bleeds.                  \\ \cmidrule{2-3}
\multicolumn{1}{c}{}                     & Predicted & \textbf{and that if you cut your finger very deeply with a knife, it usually bleeds.}                                        \\ \cmidrule{1-3} 
\multirow{2}{*}{(6)}                      & Ground Truth                & I can creep under the door, so either way I'll get into the garden, and I don't care which happens.                  \\ \cmidrule{2-3}
\multicolumn{1}{c}{}                     & Predicted & \textbf{I can creep under the door, so either way I'll get into the garden, and I don't care which happens.}                                        \\ \cmidrule{1-3} 
\multirow{2}{*}{(7)}                      & Ground Truth                & But it's no use now, thought poor Alice, to pretend to be two people while there's hardly enough of me to make one respectable person.                  \\ \cmidrule{2-3}
\multicolumn{1}{c}{}                     & Predicted & \textbf{But it's no use now, thought poor Alice, to pretend to be two people while there's hardly enough of me to make one respectable person.}                                        \\ \cmidrule{1-3}
\multirow{2}{*}{(8)}                      & Ground Truth                & She was now only ten inches high, and her face brightened up at the thought that she was now the right size for going through the little door into that lovely garden.
                  \\ \cmidrule{2-3}
\multicolumn{1}{c}{}                     & Predicted & \textbf{She was now only ten inches high, and her face brightened up at the thought that she is now the right size for going through the little door into that lovely garden.}                                        \\ \cmidrule{1-3}
\multirow{2}{*}{(9)}                      & Ground Truth                & for she had read several nice little histories about children who'd gotten burnt and eaten up by wild beasts and other unpleasant things.
                  \\ \cmidrule{2-3}
\multicolumn{1}{c}{}                     & Predicted & \textbf{for she had read several nice little histories about children who'd gotten burnt and eaten up by wild beasts and other unpleasant things.}                                        \\ \cmidrule{1-3}
\multirow{2}{*}{(10)}                      & Ground Truth                & What a curious feeling, said Alice.
                  \\ \cmidrule{2-3}
\multicolumn{1}{c}{}                     & Predicted & This time, she found a little bottle on it.                                        \\ \cmidrule{1-3}
\multirow{2}{*}{(11)}                      & Ground Truth                & Once or twice she peeped into the book her sister was reading.
                  \\ \cmidrule{2-3}
\multicolumn{1}{c}{}                     & Predicted & \textbf{Once or twice she peeped into the book her sister was reading.}                                        \\ \cmidrule{1-3}
\multirow{2}{*}{(12)}                      & Ground Truth                & how she longed to get out of that dark hall and wander about among those beds of bright flowers and those cool fountains, but she could not even get her head through the doorway.
                  \\ \cmidrule{2-3}
\multicolumn{1}{c}{}                     & Predicted & \textbf{how she longed to get out of that dark hall and wander about among those beds of bright flowers and those cool fountains, but she could not even get her head through the doorway.}                                        \\ \cmidrule{1-3}
\multirow{2}{*}{(12)}                      & Ground Truth                & Either the well was very deep, or she fell very slowly.
                  \\ \cmidrule{2-3}
\multicolumn{1}{c}{}                     & Predicted & \textbf{Either the well was very deep, or she fell very slowly.}                                        \\ \cmidrule{1-3}
\multirow{2}{*}{(13)}                      & Ground Truth                & But alas for poor Alice, when she got to the door...
                  \\ \cmidrule{2-3}
\multicolumn{1}{c}{}                     & Predicted & \textbf{But alas for poor Alice, when she got to the door...}                                        \\ \cmidrule{1-3}
\multirow{2}{*}{(14)}                      & Ground Truth                & For my end, you know, said Alice to herself, in my going out altogether like a candle.
                  \\ \cmidrule{2-3}
\multicolumn{1}{c}{}                     & Predicted & \textbf{For my end, you know, said Alice to herself, in my going out altogether like a candle.}                                        \\ \cmidrule{1-3}
\multirow{2}{*}{(15)}                      & Ground Truth                & Do you think you could manage it?
                  \\ \cmidrule{2-3}
\multicolumn{1}{c}{}                     & Predicted & \textbf{Do you think you could manage it?}                                        \\
\bottomrule
\end{tabular}
}
\centering
\caption{
Additional samples generated on \textit{Brennan} dataset. \textbf{Bold} denotes a correct match.
\label{tab:brennan_generation} }
\end{table*}

\begin{table*}[!t]
\small
\resizebox{1.0\textwidth}{!}{
\begin{tabular}{llp{14cm}}              
\toprule
{\multirow{2}{*}{(1)}} & Ground Truth                & Roy stooped to pick up a big white rock that looked like a dirty lump of chalk and handed it to Chad                                                \\ \cmidrule{2-3}
\multicolumn{1}{c}{}                     & Predicted & \textbf{Roy stooped to pick up a big white rock that looked like a dirty lump of chalk and handed it to Chad}        \\ \cmidrule{1-3} 
\multirow{2}{*}{(2)}                      & Ground Truth                & Arthur and his wine                                                             \\ \cmidrule{2-3}
\multicolumn{1}{c}{}                     & Predicted & I may finish this story                                                 \\ \cmidrule{1-3}
\multirow{2}{*}{(3)}                      & Ground Truth                & holding fidgeting conveyed glanced after sure rotting believing suppose water malignant replied                                                             \\ \cmidrule{2-3}
\multicolumn{1}{c}{}                     & Predicted & \textbf{Holding fidgeting conveyed glanced after sure rotting believing suppose water malignant replied}                             \\ \cmidrule{1-3}
\multirow{2}{*}{(4)}                      & Ground Truth                & We spent the next hour stomping around the hill while he said things like it was right here                                                             \\ \cmidrule{2-3}
\multicolumn{1}{c}{}                     & Predicted & \textbf{We spent the next hour stomping around the hill while he said things like it was right here}                             \\ \cmidrule{1-3}
\multirow{2}{*}{(5)}                      & Ground Truth                & there sounded slipped told mentioned for device issued all kentucky traffic whoever voice pushing                                                             \\ \cmidrule{2-3}
\multicolumn{1}{c}{}                     & Predicted & \textbf{There sounded slipped told mentioned for device issued all kentucky traffic whoever voice pushing}                             \\ \cmidrule{1-3}
\multirow{2}{*}{(6)}                      & Ground Truth                & Collapsing at its base Allan wrapped his arms around the stoic tree and let forth a moan a cry of purest agony that escaped him as the first tears seeped from the corners of his eyes and slid down his cheeks falling to the ground and seeping though the fallen leaves and needles to join the water of the stream flowing through the ground beneath them                                                             \\ \cmidrule{2-3}
\multicolumn{1}{c}{}                     & Predicted & \textbf{Collapsing at its base Allan wrapped his arms around the stoic tree and let forth a moan a cry of purest agony that escaped him as the first tears seeped from the corners of his eyes and slid down his cheeks falling to the ground and seeping though the fallen leaves and needles to join the water of the stream flowing through the grounds beneath them}                             \\ \cmidrule{1-3}
\multirow{2}{*}{(7)}                      & Ground Truth                & She seemed so self conscious and shallow on the outside but having that incredible gift                                       \\ \cmidrule{2-3}
\multicolumn{1}{c}{}                     & Predicted & \textbf{She seemed so self conscious and shallow on the outside but having that incredible gift}                             \\ \cmidrule{1-3}
\multirow{2}{*}{(8)}                      & Ground Truth                & It s hail across the and Tara spun to retake her seat at the helm                                       \\ \cmidrule{2-3}
\multicolumn{1}{c}{}                     & Predicted & I shall consider it in the meantime however I must be off                             \\ \cmidrule{1-3}
\multirow{2}{*}{(9)}                      & Ground Truth                & I put away the cell and used the motion to cover checking the knife in my sleeve and used one leg to check the other in my sock                                       \\ \cmidrule{2-3}
\multicolumn{1}{c}{}                     & Predicted & But I always should come now immediately before the probe is reported late                             \\ \cmidrule{1-3}
\multirow{2}{*}{(10)}                      & Ground Truth                & You could step on that marker and make the gestures the device and it would be like pushing a button in a very complex machine hu                                       \\ \cmidrule{2-3}
\multicolumn{1}{c}{}                     & Predicted & It speaks to the deepest instinct within us all yet is entirely original                             \\ \cmidrule{1-3}
\multirow{2}{*}{(11)}                      & Ground Truth                & destroyed another story last night                                       \\ \cmidrule{2-3}
\multicolumn{1}{c}{}                     & Predicted & \textbf{Destroyed another story last night}                             \\ \cmidrule{1-3}
\multirow{2}{*}{(12)}                      & Ground Truth                & Chad finished formula but this time he mind that Roy fell for it                                       \\ \cmidrule{2-3}
\multicolumn{1}{c}{}                     & Predicted & \textbf{Chad finished formula but this time he mind that Roy fell for it}                             \\ \cmidrule{1-3}
\multirow{2}{*}{(13)}                      & Ground Truth                & remote room voice truck would so what going silver taught screaming toads play being                                       \\ \cmidrule{2-3}
\multicolumn{1}{c}{}                     & Predicted & \textbf{Remote room voice truck would so what going silver taught screaming toads play being}                   \\ \cmidrule{1-3}
\multirow{2}{*}{(14)}                      & Ground Truth                & Tell them and they will create an audience                                       \\ \cmidrule{2-3}
\multicolumn{1}{c}{}                     & Predicted & \textbf{Tell them and they will create an audience}                   \\ \cmidrule{1-3}
\multirow{2}{*}{(15)}                      & Ground Truth                & Allan took a sandwich between his fingers                                       \\ \cmidrule{2-3}
\multicolumn{1}{c}{}                     & Predicted & This is the ounces which I mentioned at the restaurant                   \\
\bottomrule
\end{tabular}
}
\centering
\caption{
Additional samples generated on the \textit{GWilliams} dataset. \textbf{Bold} denotes a correct match.
\label{tab:gwilliams_generation} }
\end{table*}

\section{Dataset Splitting}
\label{sec:split}
In this section, we detail the dataset-splitting strategies employed in our study. As shown in Table~\ref{tab:dataset_split}, four distinct strategies are utilized, each presenting different levels of evaluation difficulty. The random shuffling strategy is the most basic, incorporating data from all subjects and trials into the training samples. The sentence-based strategy is more challenging, simulating scenarios where samples from different participants are not aligned, resulting in missing data for some sentences for each participant. The session-based and subject-based strategies are the most difficult but also the most realistic, as they assess the model's ability to generalize to new trials and subjects, respectively. This capability is crucial for the practical application of language-based BCIs. The \textit{Brennan} dataset utilizes only two splitting methods due to its inclusion of data from a single trial. Consequently, splitting by sentence yields results similar to those obtained by random shuffling.

\section{Implementation Details}
\label{sec:imp}
The training configurations for our model vary across different datasets and training stages. Detailed settings for each training phase are outlined in Table~\ref{tab:imp_details}. The final model is selected based on the lowest validation loss. Notably, no data augmentation techniques are employed, and no subject-related information is provided to the model.

% \section{Reconstructed Mel Spectrograms}
% \label{sec:Mel}

% \section{Generated Samples}
\section{Examples of Generated Sentences}
\label{sec:samples}
A selection of samples generated from different methods are shown in Table~\ref{tab:generated_samples}. These examples indicate that BrainECHO can produce sentences that closely match the original text, even when the reference is long and intricate. Remarkably, even without the final fine-tuning of Whisper, BrainECHO still generates results highly relevant to the original text, highlighting the effectiveness of brain-audio latent space alignment (stage 2). In contrast, EEG-to-Text~\citep{wang2022open} experiences difficulties in generating semantically relevant sentences, and NeuSpeech~\citep{yang2024decode} may generate content unrelated to the ground truth when decoding long sentences, which can have a significant impact on practical applications in high-precision decoding scenarios.

To intuitively demonstrate the powerful decoding ability of BrainECHO, additional translated examples for the \textit{Brennan} and \textit{GWilliams} datasets are presented in Table~\ref{tab:brennan_generation} and~\ref{tab:gwilliams_generation}, respectively. For most test samples, our method demonstrates accurate decoding. However, for certain samples, our model generates completely unrelated content, such as "There were doors all around the hall." and "What a curious feeling, said Alice." in Table~\ref{tab:brennan_generation}. This suggests that the model may struggle with discriminability in sentences of similar length, highlighting the persistent challenge of extracting semantically relevant patterns from low signal-to-noise non-invasive signals.

Some samples of Mel spectrograms reconstructed from the brain signals for the \textit{GWilliams} datasets are shown in Figure~\ref{fig:gwilliams_Mel}.

\begin{figure}[h]
\centering
\includegraphics[width=0.40\textwidth]{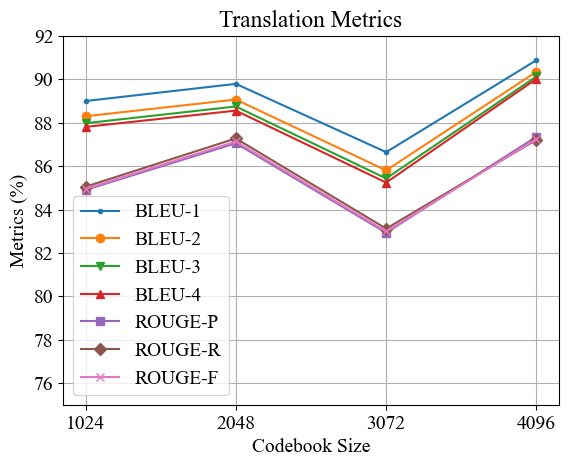} 
\caption{Translation performance using various codebook sizes on \textit{Brennan} dataset.}
\label{fig:cb_size}
\end{figure}

\section{Additional Experiments on Codebook Size in the Quantizer}    
\label{sec:codebook}
To further explore the impact of the quantizer, we investigate the performance of BrainECHO with codebook sizes ranging from 1024 to 4096. As shown in Figure~\ref{fig:cb_size}, the performance peaks at a codebook size of 4096. However, the metrics do not increase linearly with codebook size. When the codebook size increases from 1024 to 2048, the decoding performance improves, but it decreases when the size further increases to 3072. This indicates that a smaller codebook may not capture diverse acoustic representations, while a larger codebook may increase training difficulty and computational burden. Thus, we choose 2048 as the codebook size for balancing performance and efficiency.

\section{The Reason Why the Test Results Make Sense}
\label{sec:sense}
Since we split the data in paired form (i.e., "Semantic Audio -- Brain Signal evoked by the Semantic Audio"), there could be cases where the same sentence, but with different brain signals from different people (e.g., Sentence $t$ -- Brain Signal $\varepsilon_{subj1}$, Sentence $t$ -- Brain Signal $\varepsilon_{subj2}$), is included in the training or test set. Therefore, the Mel spectrograms during stage 1 could have already been seen during training, even though they are part of the test set.

However, for non-invasive natural language brain-computer interfaces, unlike invasive systems that decode neural activity related to language-specific motor areas~\citep{willett2023high, metzger2023high, chen2024neural}, non-invasive interfaces have lower signal-to-noise ratios. Furthermore, similar to invasive systems, decoding of brain signals occurs on previously seen sentences, with the vocabulary expanding progressively to achieve open vocabulary. Testing with completely unseen sentences can be overly ambitious, as demonstrated in NeuSpeech~\cite{yang2024decode}, where testing on completely unseen sentences resulted in a BLEU-1 score of only 6.91. Like the baseline in the paper, we ensure that the same sentence and its evoked brain signal do not appear in both the training and testing stages. Additionally, we have tested various data split scenarios (session, sentence and subject).

Moreover, to prevent data leakage, even if Mel spectrograms from the test set were exposed during stage 1, the brain signals from the test set were never used in stage 2. Stage 1 only serves to obtain a low-dimensional representation of the Mel spectrograms, akin to creating a feature selector for Mel spectrograms. The brain signals decoded during testing are always from data that was not seen during training.

\end{document}